\documentclass{article}

\usepackage{arxiv_template}

\usepackage[utf8]{inputenc} %
\usepackage[T1]{fontenc}    %
\usepackage{hyperref}       %
\usepackage{url}            %
\usepackage{booktabs}       %
\usepackage{amsfonts}       %
\usepackage{nicefrac}       %
\usepackage{microtype}      %
\usepackage{lipsum}
\usepackage{fancyhdr}       %
\usepackage{graphicx}       %

\usepackage{caption}
\usepackage{subcaption}
\usepackage{adjustbox}
\usepackage{multirow}
\usepackage[numbers]{natbib}
\usepackage{float}

\usepackage{amsmath}
\usepackage{amssymb}
\usepackage{mathtools}
\usepackage{amsthm}

\usepackage[capitalize,noabbrev]{cleveref}

\theoremstyle{plain}
\newtheorem{theorem}{Theorem}[section]

\theoremstyle{definition}

\newtheorem{problem}[theorem]{Problem}
\newtheorem{example}[theorem]{Example}
\theoremstyle{remark}

\usepackage{hyperref}

\title{\textbf{\centering WOODS: Benchmarks for Out-of-Distribution Generalization in Time Series}}

\author{
  Jean-Christophe Gagnon-Audet \\
  Mila - Qu\'{e}bec AI Institute\\
  Universit\'{e} of Montr\'{e}al \\
  \texttt{jean-christophe.gagnon-audet@mila.quebec} \\
  \And
  Kartik Ahuja\\
  Mila - Qu\'{e}bec AI Institute\\
  Universit\'{e} of Montr\'{e}al \\
  \And
  Mohammad-Javad Darvishi-Bayazi\\
  Mila - Qu\'{e}bec AI Institute\\
  Universit\'{e} of Montr\'{e}al \\
  \And
  Pooneh Mousavi\\
  Gina Cody School of Engineering and Computer Science\\
  Concordia University\\
  \And
  Guillaume Dumas\\
  Mila - Qu\'{e}bec AI Institute\\
  CHU Sainte-Justine Research Center, Department of Psychiatry\\
  Universit\'{e} of Montr\'{e}al \\
  \And
  Irina Rish\\
  Mila - Qu\'{e}bec AI Institute\\
  Universit\'{e} of Montr\'{e}al \\
}

\begin{document}

\maketitle

\begin{abstract}
    Deep learning models often fail to generalize well under distribution shifts. Understanding and overcoming these failures have led to a new research field on Out-of-Distribution (OOD) generalization. Despite being extensively studied for static computer vision tasks, OOD generalization has been severely underexplored for time series tasks. To shine a light on this gap, we present WOODS: 10 challenging time series benchmarks covering a diverse range of data modalities, such as videos, brain recordings, and smart device sensory signals. We revise the existing OOD generalization algorithms for time series tasks and evaluate them using our systematic framework. Our experiments show a large room for improvement for empirical risk minimization and OOD generalization algorithms on our datasets, thus underscoring the new challenges posed by time series tasks. Code and documentation are available at \url{https://woods-benchmarks.github.io/}.
\end{abstract}

\section{Introduction} \label{sect:intro}
In the last decade, the success of deep learning has led to impactful applications spanning many fields~\cite{alexnet,transformer,alphago,alphafold,gpt3}. However, parallel to this surge, there is growing evidence that deep learning models exploit undesired correlations due to selection biases, confounding factors, and other biases in the data~\cite{shortcutlearning,oodsurvey,theoretical_framework_ood}. These biases can often create shortcuts that help the model arrive at low empirical risk on a dataset. Nevertheless, a prediction rule relying on these shortcuts will not generalize out of its training distribution as it uses spuriously correlated factors instead of causal factors~\cite{causal-adaptation,causal-representation}. Such a failure becomes very concerning in real-life applications that directly impact human lives, such as medicine~\cite{DL-image-proc,DL-medecine-risk,DL-records} or self-driving cars~\cite{self-driving-review,self-driver-review2}.

\begin{figure}[h]
    \centering
    \includegraphics[width=\linewidth]{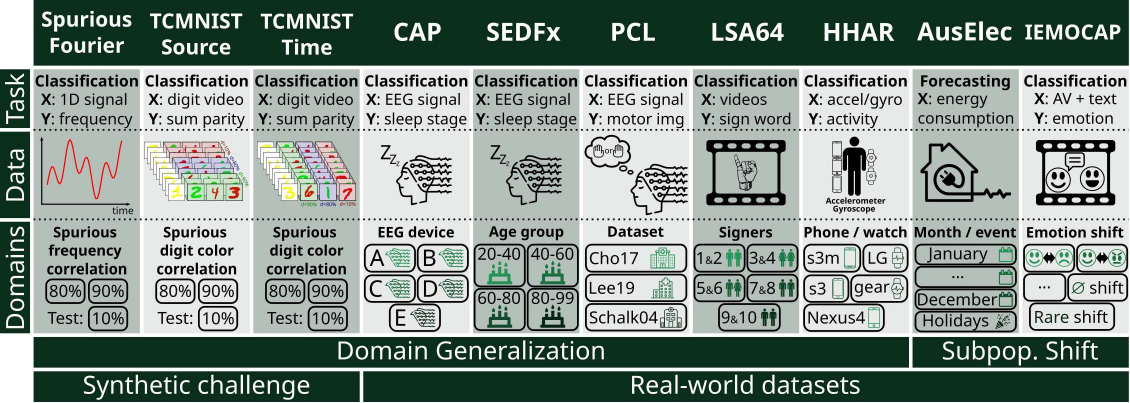}
    \caption{Summary of WOODS benchmark: tasks, modalities, domains and distribution shifts.}
    \label{fig:dataset_table}
\end{figure}

Let us explain an important failure mode with a common example from the work of \citet{terraincognita}. Consider the task of distinguishing cows and camels in pictures. The training dataset is heavily tainted by selection bias, as the vast majority of cow images were taken in green pastures, and the vast majority of camel images were taken in sandy areas. A model trained to minimize empirical risk over the training dataset leverages the selection bias and ends up using green background to classify cows and beige backgrounds to classify camels. As a way to capture different failures of deep learning models, much work has gone into finding and standardizing datasets with distribution shifts~\cite{domainbed,ood-bench,wilds}. These datasets provide a direction for research efforts in the field of OOD generalization. \citet{domainbed} gathered seven standard image datasets with distribution shifts and concluded that no OOD generalization algorithm considerably outperformed ERM, highlighting the need for better and more versatile solutions. \citet{ood-bench} showed that some algorithms outperform ERM on specific types of shifts, highlighting that different algorithms might be needed for different type of distribution shifts. \citet{wilds} created a set of benchmarks of in-the-wild distribution shifts, highlighting the challenges in real-world applications. Further related works can be found in Appendix~\ref{appendix:relatedworks}.

The above mentioned works have led to crucial empirical and theoretical insights towards addressing the OOD generalization failure in deep learning. However, they have been predominantly focused on static computer vision tasks, leaving the field of time series severely underexplored despite being essential to various applications such as computational medicine~\cite{time-series-medecine,time-series-comp-medecine,clairvoyance}, natural sciences~\cite{time-series-biology,time-series-physics}, finance~\cite{time-series-finance,time-series-finance2,time-series-finance3}, climate~\cite{time-series-climate}, retail~\cite{time-series-retail}, ecology~\cite{time-series-ecology,time-series-ecology2}, energy~\cite{time-series-building-energy} and many more~\cite{forecasting_review,forecasting-review-2}. In this work, we take the first step towards a deeper understanding of distribution shifts in time series data. Our key contributions are:

\begin{itemize}
    \item We propose WOODS: a benchmark of 3 synthetic challenge and 7 real-world datasets, totaling 10 datasets spanning a wide array of critical problems and data modalities, such as videos, brain recordings, and smart device sensory signals (See Figure~\ref{fig:dataset_table}). 
    \item We develop a systematic framework 
    for easy evaluation of new time series datasets and algorithms. The framework includes adaptation of existing OOD generalization algorithms for time series datasets.
    \item 
    We conduct extensive experiments on the above datasets with empirical risk minimization (ERM) and various OOD generalization algorithms. Our findings lead us to conclude that OOD generalization in time series brings its own set of challenges and that there is a large room for improvement as shown in Table~\ref{table:gen_gap}. 
\end{itemize}

\begin{table}[h]
    \centering
    \vspace{-13.2pt}
    \caption{Generalization gap between the In-Distribution (ID) performance and the OOD performance of ERM on the WOODS benchmarks. See Section~\ref{sect:results} for more details.}
    \begin{minipage}{\linewidth}
        \centering
        \begin{tabular}{lccc}
            \toprule
            \textbf{Dataset} & \multicolumn{2}{c}{\textbf{Performance}} &  \\
            \cmidrule{2-3}\\[-3.4ex]
            {\scriptsize (Perf. is accuracy} & \multirow{2}{*}{ID} & \multirow{2}{*}{OOD} & \multirow{2}{*}{\textbf{Gap}} \\[-0.7ex]
            {\scriptsize unless specified)} & & & \\[-0.3ex]
            \midrule
            Spur.-Fourier & $74.5\,(0.1)$ & $9.8\,(0.2)$ &  $64.7$\\
            TCM. Source & $68.4\,(0.1)$ & $10.2\,(0.1)$ &  $58.2$ \\
            TCM Time & $89.4\,(0.0)$ & $10.0\,(0.0)$ & $79.3$ \\
            \midrule
            CAP & $75.1\,(0.7)$ & $62.8\, (0.6)$ & $12.3$ \\
            SEDFx & $72.5\,(0.4)$ & $67.3\, (0.8)$ & $5.2$ \\
            PCL & $73.6\,(0.2)$ & $64.3\, (0.5)$ & $9.3$ \\
            HHAR & $93.4\,(0.4)$ & $84.4\, (0.6)$ & $9.0$ \\
            LSA64 & $86.6\,(1.0)$ & $53.4\, (2.0)$ & $33.2$ \\
            \midrule
            AusElec {\scriptsize (rmse)} & $232\,(3)$ & $397\,(9)$ & $165$ \\
            IEMOCAP & $69.1\,(0.4)$ & $57.7\,(1.9)$ & $11.4$ \\
            \bottomrule
        \end{tabular}
    \end{minipage}
    \label{table:gen_gap}
\end{table}
\newpage

\paragraph{Why OOD generalization in time series?} Recently, work in the deep learning community have shown that large scale pretrained models such as CLIP~\cite{clip} show considerable improvements when it comes to OOD generalization performance for static computer vision tasks~\cite{cha2022domain}. Since large scale pretrained models for time series data do not exist yet, whether or not large scale pretraining on time series data helps address OOD generalization challenge of time series remains to be determined. We hope our datasets and benchmarks help shed light on this important question.\\
In the next section, we discuss problem formulation, followed by discussion on the various datasets we use. In Section~\ref{sect:baselines}, we describe the  adaptation of existing methods for time series settings. In Section~\ref{sec:experiments}, we discuss the results followed by the conclusion and limitations in Section~\ref{sec: conclusion}.

\section{Problem formulation}
\label{sec: prob_formulation}
\subsection{Static tasks} \label{sect:prob_formulation_static}
Consider the standard OOD generalization setting for static supervised learning tasks. Data samples: $(X,\,Y)$ consists of the input observation $X$ and the corresponding label $Y$. We gather the datasets $D^d$ from the domains $d\in\mathcal E^\mathsf{train}$ which follow the follows the distribution $\mathbb P^{d}(X,Y)$. Datasets from these domains form the training data $D^\mathsf{train} = \cup_{d \in \mathcal{E}^{\mathsf{train}}} D^{d}$.
We define a predictor $f$. The performance of $f$ on domain $d$ is measured in terms of the risk $R^{d}(f) = \mathbb E^d\big[\ell (f(X), Y)\big]$, where $\mathbb{E}^d$ is the expectation over the distribution $\mathbb P^d$ and $\ell \rightarrow \mathbb{R}_{\geq 0}$ denotes the loss function. 
We evaluate the predictor on a set of test domains denoted as $\mathcal{E}^{\mathsf{all}}$ that could include unseen domains ($\mathcal{E}^{\mathsf{train}} \subseteq \mathcal{E}^{\mathsf{all}}$). The goal of OOD generalization is to use the training dataset $D^\mathsf{train}$ and construct a predictor $f$ that can perform well on the test domains. We write this objective formally below.

\begin{problem}\label{problem:OOD_static}
    Find a predictor $f^*$ that solves $\min_f \max_{d\in\mathcal E^\mathsf{all}} R^{d}(f)$.
\end{problem}
In the above problem, some restrictions are necessary on the set of testing domains $\mathcal E^\mathsf{all}$ to make Problem~\ref{problem:OOD_static} of practical interest. Otherwise, the best predictor is random guessing, as nothing can be assumed about the test domains. Many works~\cite{IRM,chen2021iterative,deepcoral} provide guarantees of generalizing to OOD domains by assuming that \textit{the relationship between the label and some subset of features (potentially a nonlinear transform of the observation~\cite{invariant-models-causal,erm-or-irm}) \textit{is invariant across all domains}.} We call this subset of features the \textit{invariant features}, and any other features that might be correlated with the label are called \textit{spurious features}. The predictor $f^*$ solving Problem \ref{problem:OOD_static} is said to be \textit{OOD-optimal}; $f^*$ relies on the invariant features that generalize to all domains in $\mathcal E^\mathsf{all}$~\cite{IGA}. Because the set of training domains $\mathcal E^{\mathsf{train}}$ is much smaller than the set of testing domains $\mathcal{E}^\mathsf{all}$, learning features that generalize to all test domains is a challenging task. 

\subsection{Time series tasks} \label{sect:problem_formulation_time_series}

Data samples consist of the input time series observation $\mathbf X = [X_{t}]_{t\in S_t}$, where $S_t$ is the set of time steps, and the set of labels $\mathbf Y = [Y_t]_{t\in S_p}$, where $S_p \subseteq S_t$ is the set of labeled time steps. 
The performance of the predictor $f$ is measured in terms of the risk $R^{d}(f) = \mathbb E^d\big[\ell (f(\mathbf X), \mathbf Y)\big]$, where the expectation is taken over time samples from domain $d$. We formalize the OOD generalization problem in time series as Problem~\ref{problem:OOD_static}.

In time series, similar to static tasks, the distribution shift can occur across data sources. Additionally, the distribution can also shift over time.
As a concrete real-world example of this characteristic, consider a predictor monitoring a person's health from vital signs gathered with a smart watch.

\begin{figure}[h]
    \centering
    \includegraphics[width=\linewidth]{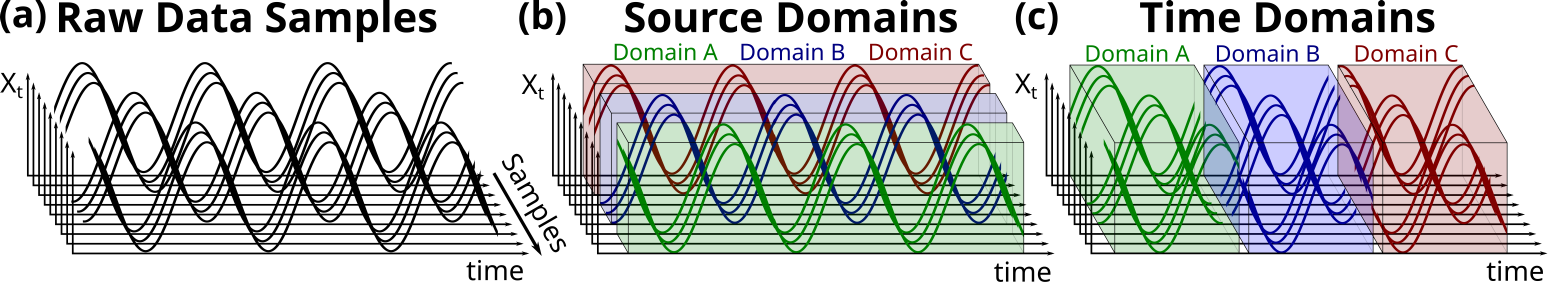}
    \caption{Illustration of the Source- and Time-domain definitions.}
    \label{fig:example_domain_definition}
\end{figure}

\begin{example}[Source-domains] \label{ex:source_domains}
    Wrist characteristics such as size or hair vary across person, or \textit{sources}. The solution to Problem~\ref{problem:OOD_static} with persons as source domains $d$ would be a predictor that does not rely on spurious wrist characteristics and thus generalizes to new persons. We call this formulation of domains as \textit{Source-domains} as time series are taken from different sources, see Figure~\ref{fig:example_domain_definition}(b).
\end{example}

\begin{example}[Time-domains] \label{ex:time_domains}
    Heart rate is lower during the night when we are asleep and higher during the day when we are awake. However, when we are working during the night, our heart rate might be higher than on a typical night. A predictor that relies on spurious features like the time of day could make a false alarm regarding our health on an atypical day.
    The solution to Problem~\ref{problem:OOD_static} with time of day as time domains $d$ would be a predictor that does not rely on spurious features, and thus generalizes to different activities at different times. We call this way of defining domains \textit{Time-domains}, as the data distribution changes through time, see Figure~\ref{fig:example_domain_definition}(c).
\end{example}

\section{Synthetic challenge datasets} \label{sect:synthetic_dataset}

\subsection{Spurious-Fourier: Spurious features encoded in the frequency domain}

\begin{figure}[h]
    \centering
    \includegraphics[width=7.455cm]{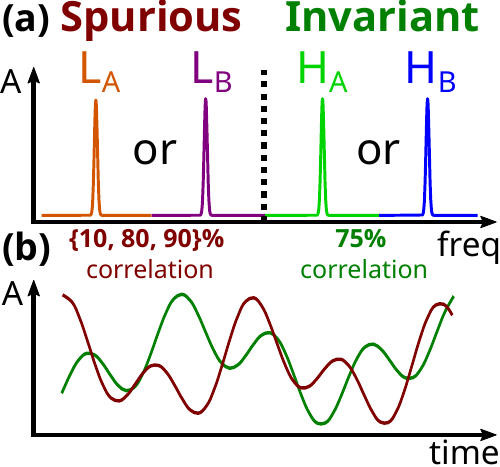}
    \caption{(a) Fourier spectrum construction in the Spurious-Fourier dataset. Signals have one low-frequency peak and one high-frequency peak. Signals are constructed from the Fourier spectrum with an inverse Fourier transform. (b) Examples of reconstructed signals, both signals have the same high frequency, but different low frequencies, which are hard to distinguish visually.}
    \label{fig:spurious_fourier_main_body}
\end{figure}

Colored MNIST (CMNIST)~\cite{IRM} presented the failure mode of ERM under distribution shift in the image domain. This was accomplished by creating training domains with strongly predictive spurious features and weakly predictive invariant features. The spurious correlation would be flipped at test time while the invariant correlation was kept the same. The correlation flip made it clear which features the model relied on to make predictions.

We create a dataset composed of one-dimensional signals, where the task is to perform binary classification based on the frequency characteristics. Signals are constructed from Fourier spectra with one low-frequency peak ($L_{A\,\mathsf{or}\,B}$) and one high-frequency peak ($H_{A\,\mathsf{or}\,B}$), see Figure \ref{fig:spurious_fourier_main_body}.Domains $D^d|_{d\in\{10\%,\,80\%,\,90\%\}}$ contain signal-label pairs, where the labels are created such that the information carried by the low-frequency signal are d\% correlated with the label (varies by domain), while the information carried by the high-frequency signal is 75\% correlated with the label.

In the training dataset $D^d|_{d\in\{80\%,\,90\%\}}$, the low-frequency signal are a stronger predictor of the label ($85\%$) than the high-frequency signal ($75\%$). Therefore, minimizing the empirical risk fails at learning the invariant high frequencies as the low frequencies achieve the lower risk.

Appendix~\ref{appendix:spurious_fourier} provides more information about the dataset.

\subsection{Temporal Colored MNIST: A study of domain definitions in sequential data}\label{sect:TCMNIST}

\begin{figure}[h]
    \centering
    \includegraphics[width=7.455cm]{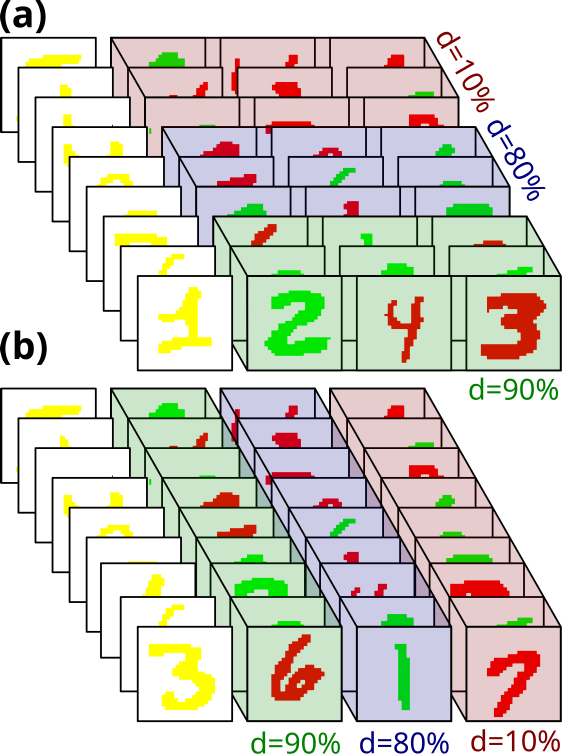}
    \caption{Domain definition of both TCMNIST (a) Source and (b) Time datasets. Data samples are videos of four colored MNIST digits where the task is to predict whether the sum of the current and previous digits in the sequence is odd or even. The color is spuriously correlated with the label.}
    \label{fig:tcmnist_main_body}
\end{figure}

In Temporal CMNIST (TCMNIST), we extend the CMNIST dataset to a binary classification task of video frames in order to investigate both domain definition paradigms presented in Section~\ref{sect:problem_formulation_time_series}: Source-domains (Example~\ref{ex:source_domains}) and Time-domains (Example~\ref{ex:time_domains}). Videos are sequences of four colored MNIST digits where the goal is to predict whether the sum of the current and previous digits in the sequence is odd or even, see Figure~\ref{fig:tcmnist_main_body}. Prediction is made for all frames except for the first one. The labels are created such that the information carried by the color of the digits are d\% correlated with the label (varies by domain), while the information carried by the value of the digit is 75\% correlated with the label.

\paragraph{TCMNIST-Source} \label{sect:TCMNIST-Source}
Domains are created such that the color correlation is constant among the frames of a video, but varies between video from different domains $d\in\{10\%,\,80\%,\,90\%\}$. The domain definition is depicted in Figure~\ref{fig:tcmnist_main_body}(a).

\paragraph{TCMNIST-Time} \label{sect:TCMNIST-Time}
Domains are created such that the color correlation varies across frames. However, videos all have the same sequence of color correlation, where the first labeled frame correlation is $90\%$, second is $80\%$ and third is $10\%$. The domain definition is depicted in Figure~\ref{fig:tcmnist_main_body}(b).

Appendix~\ref{appendix:tcmnist} provides more information about the dataset.

\section{Real-world datasets} \label{sect:woods_datasets}

\subsection{CAP: Sleep classification across different machines}

\begin{figure}[h]
    \centering
    \includegraphics[width=7.455cm]{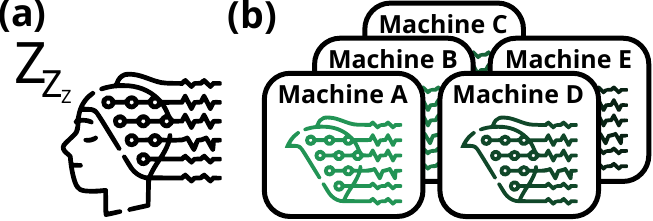}
    \caption{Summary of the CAP dataset. (a) The task is to perform sleep stage classification from EEG measurements. (b) The dataset has five source domains, where each domain contains data gathered with a different machine. The goal is to generalize to unseen machines.}
    \label{fig:cap_main_body}
\end{figure}

A recurrent problem in computational medicine is that models trained on data from a given recording device will not generalize to data coming from another device, even when both devices are from a similar equipment provider. Failure to generalize to unseen machines 
can cause critical issues for clinical practice because a false sense of confidence in a model could lead to a false diagnosis~\cite{repro_in_bioinformatics,robust-EEG}. We study these machinery-induced distribution shifts with the CAP~\cite{cap,physionet} dataset (Figure~\ref{fig:cap_main_body}).

We consider the sleep stage classification task from electroencephalographic (EEG) measurements. The dataset has five source domains, where each domain contains data gathered with a different machine. The goal is to generalize to unseen machines.

Appendix~\ref{appendix:cap} provides more information about the dataset.

\subsection{SEDFx: Sleep classification across age groups}

\begin{figure}[h]
    \centering
    \includegraphics[width=7.455cm]{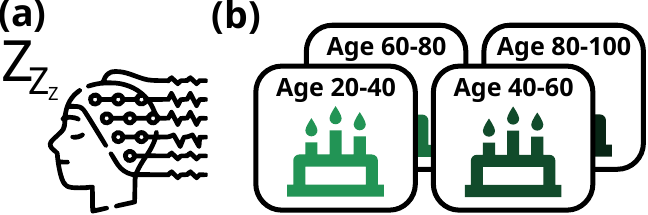}
    \caption{Summary of the SEDFx dataset. (a) The task is to perform sleep stage classification from EEG measurements. (b) The dataset has four source domains, where each domain contains data from participants of a certain age group. The goal is to generalize to unseen age groups.}
    \label{fig:sedfx_main_body}
\end{figure}

In clinical settings, we train a model on the data gathered from a limited number of patients and hope this model will generalize to new patients in the future~\cite{worstgroup_subpopulation}. However, this generalization between observed patients in the training dataset and new patients is not guaranteed. Distribution shifts caused by shifts in patient demographics (e.g., age, gender, and ethnicity) can cause the model to fail. We study age demographic shift with the SEDFx~\cite{sedfx,physionet} dataset (Figure~\ref{fig:sedfx_main_body}). 

We consider the sleep classification task from EEG measurements. The dataset has four source domains, where each domain contains data from participants of a certain age group. The goal is to generalize to unseen age groups.

Appendix~\ref{appendix:sedfx} provides more information about the dataset.

\subsection{PCL: Motor imagery classification across data-gathering procedures}

\begin{figure}[h]
    \centering
    \includegraphics[width=7.455cm]{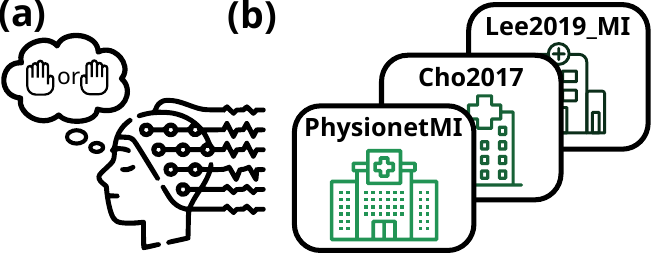}
    \caption{Summary of the PCL dataset. (a) The task is to perform motor imagery classification from EEG measurements. (b) The dataset has three source domains, where each domain contains a dataset from a different research group carrying out the same task. The goal is to generalize to unseen datasets of the same task.}
    \label{fig:pcl_main_body}
\end{figure}

Aside from changes in the recording device and shifts in patient demographics, human intervention in the data gathering process is another contributing factor to the distribution shift that can lead to failure of clinical models (e.g., Camelyon17~\cite{wilds,WILDS-v2}). This challenge is especially prevalent in temporal medical data (e.g., EEG, MEG, and others) because recording devices are complex tools greatly affected by nonlinear effects and modulations. These effects are often caused by context and preparations made before the recording~\cite{robust-EEG}. We study these procedural shifts with the PCL~\cite{lee2019,cho2017,physionetMI,moabb} dataset (Figure~\ref{fig:pcl_main_body}).

We consider the motor imagery task from EEG measurements. The dataset has three source domains, where each domain contains a dataset from a different research group carrying out the same task. The goal is to generalize to unseen datasets of the same task.

Appendix~\ref{appendix:pcl} provides more information about the dataset.

\subsection{LSA64: Sign language video classification across speakers}

\begin{figure}[h]
    \centering
    \includegraphics[width=7.455cm]{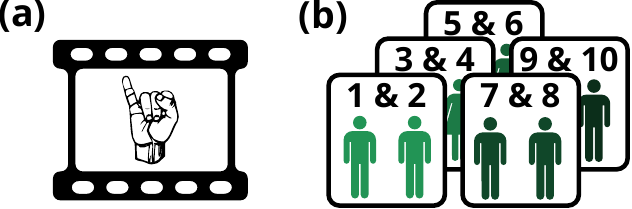}
    \caption{Summary of the LSA64 dataset. (a) The task is to perform signed word classification from videos. (b) The dataset has five source domains, where each domain contains videos of different signers. The goal is to generalize to unseen signers.}
    \label{fig:lsa64_main_body}
\end{figure}

Communication is an individualistic way to convey information through different media: text, speech, body language, and many others. However, some media are more distinctive and challenging than others. For example, text communication has less inter-individual variability than body language or speech. If deep learning systems hope to interact with humans effectively, models need to generalize to new and evolving mannerisms, accents, and other subtle variations in communication that significantly impact the meaning of the message conveyed. We study the ability of models to recognize information coming from unseen individuals with the LSA64~\cite{lsa64} dataset (Figure~\ref{fig:lsa64_main_body}).

We consider the video classification of signed words in Argentinian Sign Language. The dataset has five source domains, where each domain contains videos of different signers. The goal is to generalize to unseen signers.

Appendix~\ref{appendix:lsa64} provides more information about the dataset.

\subsection{HHAR: Human activity recognition across smart devices}

\begin{figure}[h]
    \centering
    \includegraphics[width=7.455cm]{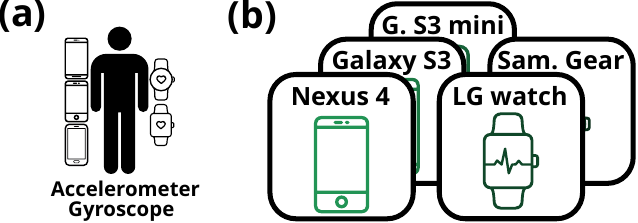}
    \caption{Summary of the HHAR dataset. (a) The task is to perform human activity classification from smart devices sensory data. (b) The dataset has five source domains, where each domain contains data gathered with a different smart device. The goal is to generalize to unseen smart devices.}
    \label{fig:hhar_main_body}
\end{figure}

The intrinsic biases from inaccurate and poorly calibrated sensors of smart devices, along with the accumulated biases from everyday use makes human activity recognition a notoriously difficult task when task when done across devices~\cite{hhar,blunck2013heterogeneity}. Contrary to static tasks where uninformative features can often be segmented out from the input features (e.g., background when classifying an animal from an image), invariant features in time series are often highly convoluted with other spurious features. We study the ability of models to ignore spurious information from complex signals with the HHAR~\cite{hhar,uci} dataset (Figure~\ref{fig:hhar_main_body}).

We consider the human activity classification task from accelerometer and gyroscope measurements of smartphones and smartwatches. The dataset has five source domains, where each domain contains data gathered with a different device. The goal is to generalize to unseen smart devices.

Appendix~\ref{appendix:hhar} provides more information about the dataset.

\subsection{AusElec: Forecasting of energy consumption throughout the year}

\begin{figure}[h]
    \centering
    \includegraphics[width=7.455cm]{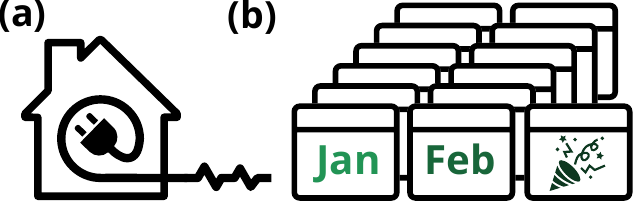}
    \caption{Summary of the AusElec dataset. (a) The task is to forecast electricity consumption. (b) The dataset has 13 time domains, where each domain contains data from different months and holidays. The goal is to perform well on all seasonalities.}
    \label{fig:aus_elec_main_body}
\end{figure}

Seasonality is the property of time series where recurring characteristics appear every cycle of a fixed period, e.g., weekly. A common practice in the forecasting field is to provide models with additional information, e.g., day of week in order to allow models to leverage seasonality for better predictions. However, holidays is a seasonality of time series that is very sparse which models often fail to capture. We study the performance of models on sparse seasonality with the AusElec~\cite{auselec,monash} dataset (Figure~\ref{fig:aus_elec_main_body})

We consider the electricity consumption forecasting task. The dataset has 13 time domains, where each domain contains data from different months and holidays. The goal is to perform well on all seasonalities. 

Appendix~\ref{appendix:auselec} provides more information about the dataset.

\subsection{IEMOCAP: Emotion recognition across different conversational emotion shifts }

\begin{figure}[h]
    \centering
    \includegraphics[width=7.455cm]{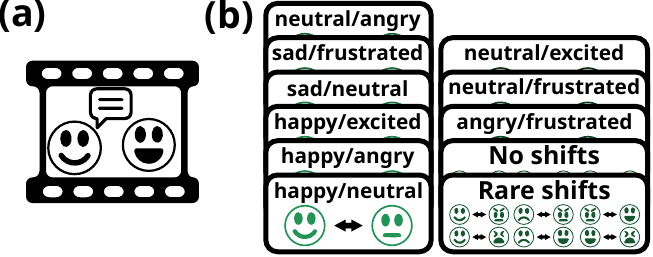}
    \caption{Summary of the IEMOCAP dataset. (a) The task is to perform emotion recognition from multi modal data (video, sound, text). (b) The dataset has 11 time domains, where each domain contains data from a different emotion shifts during conversations. The goal is to perform well on all conversational emotion shifts.}
    \label{fig:iemocap_main_body}
\end{figure}

Speakers tend to maintain an emotional state over a conversation. However, external stimuli can invoke a shift in the emotional state of speakers~\cite{poria2019emotion}. Such emotion shift are often sparsely represented in the data, making it hard for models to classify them adequately.
Recent work on emotion recognition models~\cite{poria2019emotion,poria2018meld,majumder2019dialoguernn} show the failure of existing models to adapt to those emotion shift. We study the performance of models on emotional shift with the IEMOCAP~\cite{busso2008iemocap} dataset (Figure~\ref{fig:iemocap_main_body}).

We consider the emotion recognition task. The dataset has 11 time domains, where each domain contains data from a different emotion shift during conversations. The goal is to perform well on all conversational emotion shifts.

Appendix~\ref{appendix:iemocap} provides more information about the dataset.

\section{Adaptation of OOD generalization algorithms to time series} \label{sect:baselines}

Many algorithms were proposed to address the failure of machine learning models under distribution shifts. However, they were formulated for the image domain and require adaptation to be used with time series. We now describe how we adapt them to the time series settings.

On top of Empirical Risk Minimization (\textbf{ERM},~\citet{ERM}), we have selected commonly used algorithms from the OOD generalization research field to adapt and evaluate on WOODS benchmarks: Invariant Risk Minimization (\textbf{IRM},~\citet{IRM}), Group Distributionally Robust Optimization (\textbf{GroupDRO},~\citet{groupdro}), Variance Risk Extrapolation (\textbf{VREx},~\citet{VREx}), Spectral Decoupling (\textbf{SD},~\citet{SD}), Information Bottleneck Empirical Risk Minimization (\textbf{IB-ERM},~\citet{IB}), Transfer (\textbf{Transfer},~\citet{transferalgorithm}), Contrastive Adversarial Domain bottleneck (\textbf{CAD},~\citet{CAD}), Conditional CAD (\textbf{CondCAD},~\citet{CAD}).

The loss function of above algorithms (except GroupDRO and Transfer) comprises of two terms: the empirical risk for a domain $R^d(f)$ and a penalty function $P(f)$. For the empirical risk of domain $d$, we average the risk across the set of labeled time steps of a time series belonging to domain $d$: $S_p^d$.
\begin{equation} \label{eq:emp_risk}
    R^d(f) = \frac{1}{n^d}\sum_{(\mathbf X, \mathbf Y)\in D} \frac{1}{|S_p^d|}\sum_{t\in S_p^d} \mathcal L\big(f(X_{1:t}),Y_t\big)
\end{equation}
where $n^d$ is the number of samples from domain $d$ in the dataset $D$. In the case of Source-domains, all time steps of a time series belongs to the same domain, while for the Time-domains there can be time steps belonging to different domains in the time series. IRM and VREx use a penalty that relies on the risk across domains, we use the risk from Equation (\ref{eq:emp_risk}) in the corresponding penalties.
\begin{equation} \label{eq:pen_val}
    P(f) = \frac{1}{n^d} \sum_{(\mathbf X, \mathbf Y)\in D} \frac{1}{|S_p^d|}\sum_{t\in S_p^d} \tilde P(f,X_{1:t}, Y_t),
\end{equation}
where $\tilde P$ is the penalty applied at each prediction point, e.g., $\tilde P(f, X_{1:t}, Y) = \|f(X_{1:t})\|^2$ for SD. 
Equations~(\ref{eq:emp_risk}) and (\ref{eq:pen_val}) are a simplifications of the adaptation; in Appendix~\ref{appendix:adaptation} we provide a more general formulation along with explicit penalty definitions for all algorithms used in this work.

\section{Experiments} \label{sec:experiments}

Our framework follows the DomainBed~\cite{domainbed} workflow for hyperparameter search and model selection for a fair and systematic evaluation of OOD generalization algorithms. We perform a random search over 20 hyperparameter configurations, which we repeat three times for error estimation. We then report the performance of the model chosen with our model selection methods (see Section~\ref{sect:model_selection}).

Appendix~\ref{appendix:framework} provides more information on the on the framework along with hyperparameter search spaces and Appendix~\ref{appendix:datasets} provides the specific architectures used for each dataset.

\subsection{Model selection methods} \label{sect:model_selection}
\paragraph{For domain generalization} We split all the training domains into training and validation sets. With \textit{train-domain validation}, we choose the model that gets the best average validation performance across training domains. With \textit{test-domain validation}, we choose the model with the best performance on the test domain, however, we restrict the test domain queries to the final training checkpoint only, effectively disallowing early stopping. With \textit{oracle train-domain validation}, we choose the model with the best performance on the test domain, however, we restrict the test domain queries to the training checkpoint with the best performance on the validation set of the training domains.
\paragraph{For subpopulation shift} We split all domains into training, validation and test sets. With \textit{domain-average validation}, we choose the model with the best average validation performance across domains. With \textit{worst-domain validation}, we choose the model with the best worst domain performance.

Appendix~\ref{appendix:model_selection} provides more details on the model selection methods, and why we chose them.

\subsection{OOD generalization algorithms results} \label{sect:results}

\paragraph{WOODS datasets have a significant generalization gap}
Table~\ref{table:gen_gap} summarizes the generalization gap for all WOODS datasets, along with the In-Distribution (ID) and OOD performance. We compute the generalization gap to be an upper bound of the attainable performance on the test domains. This is positively indicative that there is significant improvements to be made over ERM. 

Appendix~\ref{appendix:measuring_ood_gap} provide more details on how the generalization gaps are obtained.

\paragraph{Marginal improvement over ERM on WOODS real-world datasets}
Table~\ref{table:real_world_results_summary} and~\ref{table:subpopulation_shift_result_summary} summarizes the baseline results on our real-world datasets\footnote{Performance of SD, IRM, CAD, CondCAD, and Transfer are not reported on forecasting datasets, because their adaptation to a forecasting task is not possible without significant alterations to their formulations.}. We observe a marginal improvement over ERM on several datasets with the adapted algorithms.
\begin{table}
    \centering    
        \caption{Summary of OOD generalization algorithms performance on the real-world datasets.}
        \begin{center}
            \adjustbox{max width=0.7\textwidth}{%
            \begin{tabular}{lccccccccc}
            \toprule
            \multicolumn{7}{c}{\textbf{Train-domain validation}} \\
            \midrule
            \multirow{2}{*}{\textbf{Objective}} & CAP & SEDFx & PCL & LSA64 & HHAR & \multirow{2}{*}{\textbf{Average}}\\[-0.7ex]
            &  {\scriptsize (accuracy)}   & {\scriptsize (accuracy)}  & {\scriptsize (accuracy)}  & {\scriptsize (accuracy)}  & {\scriptsize (accuracy)}  \\[-0.1ex]
            \midrule
            ERM &  $62.8 \,(0.6)$  &  $67.3 \,(0.8)$  &  $64.3 \,(0.5)$  &  $53.4 \,(2.0)$  &  $84.4 \,(0.6)$  &  66.4 \\
            IRM &  $58.7 \,(1.3)$  &  $62.7 \,(0.7)$  &  $63.9 \,(0.2)$  &  $45.0 \,(1.6)$  &  $82.9 \,(0.9)$  &  62.6 \\
            VREx &  $48.6 \,(1.7)$  &  $56.1 \,(1.4)$  &  $63.2 \,(0.3)$  &  $46.8 \,(2.9)$  &  $83.2 \,(0.5)$  &  59.6 \\
            GroupDRO &  $62.0 \,(0.8)$  &  $65.2 \,(0.8)$  &  $\mathbf{64.8} \,(0.3)$  &  $46.3 \,(2.1)$  &  $84.2 \,(0.4)$  &  64.5 \\
            IB-ERM &  $\mathbf{63.2} \,(0.8)$  &  $69.5 \,(0.5)$  &  $64.4 \,(0.3)$  &  $\mathbf{57.3} \,(1.9)$  &  $83.5 \,(0.7)$  &  \textbf{67.6} \\
            SD &  $60.8 \,(0.9)$  &  $\mathbf{69.8} \,(0.5)$    &  $64.4 \,(0.2)$  &  $50.7 \,(1.7)$  &  $\mathbf{85.6} \,(0.1)$  &  66.2 \\
             CAD &  $62.2 \, (1.1)$  &  $66.1 \, (0.5)$  &  $64.6 \, (0.6)$  &  $50.3 \, (2.2)$  &  $85.0 \, (0.6)$  &  65.6 \\
             CondCAD  & $62.6 \, (0.5)$  &  $66.1 \, (0.8)$  &  $64.2 \, (0.3)$  &  $53.4 \, (1.5)$  &  $84.3 \, (0.8)$  &  66.1 \\
             Transfer  & $55.0\, (1.3)$  &  $61.0\, (0.9)$  &  $62.3\, (0.2)$  &  $47.3\, (1.3)$  &  $84.4\, (0.5)$  &  62.0 \\
            \bottomrule
            \end{tabular}}
            \end{center}
            \begin{center}
            \adjustbox{max width=0.7\textwidth}{%
            \begin{tabular}{lcccccc}
                \toprule
                \multicolumn{7}{c}{\textbf{Oracle train-domain validation}} \\
                \midrule
                \multirow{2}{*}{\textbf{Objective}} & CAP & SEDFx & PCL & LSA64 & HHAR & \multirow{2}{*}{\textbf{Average}}\\[-0.7ex]
                &  {\scriptsize (accuracy)}   & {\scriptsize (accuracy)}  & {\scriptsize (accuracy)}  & {\scriptsize (accuracy)}  & {\scriptsize (accuracy)}  \\[-0.1ex]
                \midrule
                ERM &  $64.2 \,(0.6)$  &  $68.5 \,(0.3)$  &  $\mathbf{65.3} \,(0.3)$  &  $58.2 \,(0.9)$  &  $85.3 \,(0.5)$  &  68.3 \\
                IRM &  $60.5 \,(0.9)$  &  $64.3 \,(0.6)$  &  $64.4 \,(0.4)$  &  $43.6 \,(2.0)$  &  $83.4 \,(0.6)$  &  63.2 \\
                VREx &  $49.3 \,(1.6)$  &  $57.0 \,(0.7)$  &  $63.3 \,(0.3)$  &  $50.0 \,(0.8)$  &  $83.2 \,(0.6)$  &  60.6 \\
                GroupDRO &  $62.9 \,(0.6)$  &  $66.1 \,(0.5)$  &  $64.5 \,(0.3)$  &  $54.0 \,(1.3)$  &  $84.3 \,(0.4)$  &  66.4 \\
                IB-ERM &  $\mathbf{65.2} \,(0.6)$  &  $\mathbf{70.6} \,(0.4)$    &  $65.0 \,(0.3)$  &  $\mathbf{59.8} \,(1.0)$  &  $85.5 \,(0.2)$  &  \textbf{69.2} \\
                SD &  $63.2 \,(0.4)$  &  $\mathbf{70.6} \,(0.4)$    &  $\mathbf{65.3} \,(0.12)$  &  $58.6 \,(1.0)$  &  $\mathbf{86.3} \,(0.2)$  &  68.8 \\
                CAD  &  $63.6 \, (0.8)$  &  $67.5 \, (0.3)$  &  $64.5 \, (0.4)$  &  $57.8 \, (1.5)$  &  $84.8 \, (0.3)$  &  67.7 \\
                CondCAD  & $63.3 \, (0.7)$  &  $66.6 \, (0.6)$  &  $63.6 \, (0.3)$  &  $57.4 \, (1.3)$  &  $84.7 \, (0.6)$  &  67.1 \\
                Transfer  & $57.7 \, (0.8)$  &  $61.5 \, (0.7)$  &  $61.9 \, (0.2)$  &  $51.3 \, (1.6)$  &  $85.1 \, (0.3)$  &  63.5 \\
            \bottomrule
            \end{tabular}}
        \end{center}
        \label{table:real_world_results_summary}
\end{table}
\begin{table}
        \caption{Summary of OOD generalization algorithms performance on the synthetic challenge datasets.}
        \begin{center}
            \adjustbox{max width=0.7\textwidth}{%
            \begin{tabular}{lccccccccc}
            \toprule
            \multicolumn{5}{c}{\textbf{Train-domain validation}} \\
            \midrule
            \multirow{2}{*}{\textbf{Objective}} & Spur.-Fourier & TCMNIST-Source & TCMNIST-Time & \multirow{2}{*}{\textbf{Average}}\\[-0.7ex]
            &  {\scriptsize (accuracy)}   & {\scriptsize (accuracy)}  & {\scriptsize (accuracy)}  &  \\[-0.1ex]
            \midrule
            ERM &  $9.9\,(0.1)$ &  $10.1\,(0.0)$ &  $9.9\,(0.1)$ &  10.0 \\
            IRM &  $10.3\,(0.1)$ &  $9.8\,(0.1)$ &  $10.3\,(0.1)$ &  10.1 \\
            VREx &  $10.4\,(0.2)$ &  $9.8\,(0.2)$ &  $9.8\,(0.1)$ &  10.0 \\
            GroupDRO &  $10.1\,(0.2)$ &  $10.4\,(0.0)$ &  $10.1\,(0.1)$ &  10.2 \\
            IB-ERM &  $9.2\,(0.3)$ &  $10.0\,(0.1)$ &  $10.2\,(0.0)$ &  9.8 \\
            SD &  $9.7\,(0.2)$ &  $10.2\,(0.1)$ &  $10.2\,(0.1)$ &  10.0 \\
            CAD &  $10.3 \, (0.4)$  &  $9.8 \, (0.1)$  &  $9.9 \, (0.1)$ & {10.0}\\
            CondCAD & $10.3 \, (0.6)$  &  $10.1 \, (0.1)$  &  $9.9 \, (0.2)$ & {10.1}\\
            Transfer & $9.5 \, (0.1)$  &  $10.0 \, (0.2)$  &  $9.8 \, (0.0)$ &{9.8} \\
            \bottomrule
            \end{tabular}}
        \end{center}
        
        \begin{center}
            \adjustbox{max width=0.7\textwidth}{%
            \begin{tabular}{lcccc}
            \toprule
            \multicolumn{5}{c}{\textbf{Test-domain validation}} \\
            \midrule
            \multirow{2}{*}{\textbf{Objective}} & Spur.-Fourier & TCMNIST-Source & TCMNIST-Time & \multirow{2}{*}{\textbf{Average}}\\[-0.7ex]
            &  {\scriptsize (accuracy)}   & {\scriptsize (accuracy)}  & {\scriptsize (accuracy)}  &  \\[-0.1ex]
            \midrule
            ERM &  $12.1\,(2.0)$ &  $30.3\,(0.8)$ &  $28.6\,(2.4)$ &  23.7 \\
            IRM &  $58.8\,(2.0)$ &  $\mathbf{52.7}\,(0.6)$ &  $\mathbf{50.6}\,(0.2)$ &  54.1 \\
            VREx &  $\mathbf{63.7}\,(0.7)$ &  $49.7\,(0.2)$ &  $\mathbf{50.6}\,(0.6)$ &  \textbf{54.7} \\
            GroupDRO &  $21.5\,(2.1)$ &  $33.5\,(2.9)$ &  $24.8\,(3.9)$ &  26.6 \\
            IB-ERM &  $18.6\,(4.0)$ &  $28.1\,(1.1)$ &  $33.7\,(6.5)$ &  26.8 \\
            SD &  $10.0\,(0.1)$ &  $27.4\,(3.5)$ &  $31.8\,(5.1)$ &  23.0 \\
            { CAD } &  $20.4 \, (2.4)$  &  $26.3 \, (3.3)$  &  $27.2 \, (2.3)$  &  {28.0} \\
            { CondCAD } & $16.0 \, (1.6)$  &  $20.6 \, (2.5)$  &  $22.7 \, (2.9)$  &  {20.1} \\
            { Transfer } & $13.4 \, (2.8)$  &  $18.3 \, (0.8)$  &  $24.2 \, (5.3)$  &  {21.6} \\
            \bottomrule
            \end{tabular}}
        \end{center}

    \label{table:Synthetic-results}
\end{table}

\begin{table}
    \centering    
    \caption{Summary of OOD generalization algorithms performance on subpopulation shifts datasets.}
    \begin{minipage}{0.49\linewidth}
        \centering
        \adjustbox{max width=\textwidth,center}{%
        \begin{tabular}{lccc}
        \toprule
        \multicolumn{3}{c}{\textbf{Domain-average validation}} \\
        \midrule
        \multirow{2}{*}{\textbf{Objective}} & AusElec &  IEMOCAP \\[-0.7ex]
          &  {\scriptsize (rsme)}  & {\scriptsize (accuracy)}  \\[-0.1ex]
        \midrule
        ERM &  $397 \,(9)$  &  $57.7 \,(1.9)$ \\
        IRM &  X  &  $55.9 \,(1.2)$  \\
        VREx &  $415 \,(10)$  &  $59.4 \,(1.4)$  \\
        GroupDRO &  $409 \,(2)$  &  $56.1 \,(1.2)$  \\
        IB-ERM &  $\textbf{394} \,(2)$  &  $\textbf{59.9} \,(0.5)$  \\
        SD &  X  &  $58.0 \,(0.4)$  \\
        \bottomrule
        \end{tabular}}
    \end{minipage}
    \begin{minipage}{0.49\linewidth}
        \adjustbox{max width=\textwidth,center}{%
        \begin{tabular}{lccc}
        \toprule
        \multicolumn{3}{c}{\textbf{Worst-domain validation}} \\
        \midrule
        \multirow{2}{*}{\textbf{Objective}}  & AusElec &  IEMOCAP \\[-0.7ex]
          &  {\scriptsize (rsme)}  & {\scriptsize (accuracy)}  \\[-0.1ex]
        \midrule
        ERM &  $404 \,(7)$  &  $56.3 \,(2.8)$ \\
        IRM &  X  &  $\textbf{58.9} \,(1.1)$ \\
        VREx &  $409 \,(4)$  &  $57.7 \,(3.1)$ \\
        GroupDRO &  $424 \,(13)$  &  $58.8 \,(1.0)$ \\
        IB-ERM &  $\textbf{391} \,(5)$  &  $58.8 \,(1.5)$ \\
        SD &  X  &  $56.1 \,(1.2)$ \\
        \bottomrule
        \end{tabular}}
    \end{minipage}
    \label{table:subpopulation_shift_result_summary}
\end{table}

\paragraph{Algorithms fail on synthetic challenge dataset with train-domain validation}
Table~\ref{table:Synthetic-results} summarizes the baseline results on our synthetic challenge datasets. We observe that IRM and VREx significantly outperform ERM on WOODS synthetic datasets with test-domain validation. However, all algorithms fail with train-domain validation as chosen models learned to rely on the spurious features which were anti-correlated with the label during testing. This caused accuracies of 10\%, significantly below the random guessing accuracy of 50\%. We also observe that IRM and VREx under perform on real-world dataset.

\section{Conclusion \& Limitations}\label{sec: conclusion}

This work introduced WOODS: a benchmark of 10 datasets for OOD generalization in time series. We formulated the Source- and Time-domain settings for dealing with different scenarios of distribution shifts in time series. We adapted OOD algorithms to the time series setting, and provided their performance on WOODS datasets using our fair and systematic evaluation framework. With WOODS, we take the first step and lay the groundwork towards understanding and solving distribution shifts failure mode of deep learning in time series.

While this work proposes an initial set of benchmarks for OOD generalization in time series, our benchmarks are inherently biased toward Source-domains problems, classification tasks, and neurophysiology modalities. We hope for WOODS to be a platform to continue building towards a complete set of benchmarks with datasets covering those missing settings and other data modalities not currently studied in WOODS.

\subsubsection*{Broader Impact Statement}

Failures of deep learning models under distribution shifts are very concerning in real-life applications that directly impact human lives, such as medicine or self-driving cars. The WOODS benchmark hopes to give researchers and engineers a meaningful measure of generalization performance to test new algorithms and alleviate potentially dangerous failures. However, it is possible that our benchmark does not accurately reflect OOD generalization performance for all possible applications. This could lead to false confidence in a deployed system that could be dangerous to human life.

\newpage

\bibliographystyle{plainnat}  
\bibliography{references}

\newpage
\appendix
\onecolumn

\section{Organization}
In Appendix~\ref{appendix:relatedworks}, we provide additional related works, including works in both OOD generalization algorithms and existing datasets in the field. In Appendix~\ref{appendix:datasets}, we provide further details on all WOODS datasets, along with model architecture choices and licenses. In Appendix~\ref{appendix:adaptation}, we provide a general formulation for OOD generalization algorithms adaptation to time series, along with explicit penalty value function definitions for the algorithms used in this work. In Appendix~\ref{appendix:measuring_ood_gap}, we give further details on how we define the generalization gaps in our datasets. In Appendix~\ref{appendix:framework}, we describe our evaluation framework. In Appendix~\ref{appendix:model_selection}, we discuss the model selection strategies used in this work.

\section{Related works} \label{appendix:relatedworks}
In the main text, we covered important benchmarks in the field of OOD generalization. In this section, we detail a broader horizon of datasets in the field along with OOD generalization algorithms.

\subsection{OOD generalization algorithms}
Several algorithms were recently proposed to address the OOD generalization failures of deep learning~\cite{IRM,VREx,SD,IB,groupdro,ANDMask,sandmask,IGA,MBDG,CAD,fishr,irmgames,TRM,ICM,JTT,nonlinIRM,DARE,GOOD,velodrome}. Several of these algorithms adopt the invariance principle from causality \cite{pearl2009causality, pearl1995causal,peters2016causal} to create predictors that rely on the causes of the label to make predictions. Invariance is leveraged because it is a more flexible and scalable alternative to conditional independence testing typically used for causal discovery~\cite{CIT,approx-CIT}. An optimal predictor that relies on the cause will be min-max optimal~\cite{erm-or-irm, ICM,invariant-models-causal} under a large class of distribution shifts. Some works have also been proposed to address the distribution shift that arises through time in time series forecasting tasks~\cite{adarnn,adaptiveairpolution,relationshipaligned}.

\subsection{Existing benchmarks for OOD generalization}

\paragraph{Synthetic datasets} Many synthetic and semi-synthetic datasets were created to gain a better understanding of generalization failure in deep learning, e.g., CMNIST~\cite{IRM} investigates our motivating cow or camel classification problem, RMNIST~\cite{rmnist} investigates invariance with respect to rotation of images, and Invariance Unit Tests~\cite{lineartest} investigates six different types of distribution shifts for linear models. 

\paragraph{Image datasets} Many real (i.e., non-synthetic) image datasets were proposed, some with naturally occurring distribution shifts and some with artificially induced distribution shifts. Several of these datasets are composed of different renditions of the same underlying labels, e.g., PACS~\cite{pacs} (Photo, Art, Cartoon, Sketch), DomainNet~\cite{domain-net} (Clipart, Infographic, Painting, Quickdraw, Photo, Sketch), Office-Home~\cite{office-home} (Art, Clipart, Product, Photo), and ImageNet-R~\cite{imagenet-R} (art, cartoons, graffiti, embroidery). Others focus on the generalization across different datasets with same rendition, such as many altered versions of ImageNet, e.g., ImageNet-A~\cite{imagenet-A} comprises of ImageNet images that are missclassified by ResNet models, ImageNet-C~\cite{imagenet-C} comprises algorithmically corrupted images from the original ImageNet, ImageNet-Sketch~\cite{imagenet-sketch} comprises samples through Google Image queries, ImageNet-V2~\cite{imagenet-v2} comprises similar images to ImageNet collected by closely following the original labeling protocol, BREEDS~\cite{breeds} comprises of ImageNet subclasses that are held out during training. Others created datasets of similar renditions but different sources, e.g., VLCS~\cite{vlcs} comprises images from four different photo datasets, ObjectNet~\cite{objectnet} comprises images from different predefined viewpoints, Terra Incognita~\cite{terraincognita} comprises images from multiple different traps. Another dataset class has strong spurious features that create shortcuts to minimize the empirical risk, e.g., in CelebA~\cite{celeba} hair color as a spurious attribute to a gender classification task, while in NICO~\cite{nicodataset}, Waterbirds~\cite{groupdro} and backgrounds challenge~\cite{backgroundchallenge} use the background as a spurious attribute of animal classification task. Finally, some other datasets were created to study specific problems, e.g., Shift15m~\cite{shift15m} that looks at OOD generalization in the large data regime.

\paragraph{Language datasets} 
Natural language is prone to distribution shifts because of interindividual variability, consequently, many works investigated OOD generalization in language. The Machine Translation dataset from the work of \citet{shifts} investigates generalization to atypical language usage in a translation task. \citet{devilisinthedetail} explored the systematic generalization of transformers with five datasets, i.e., SCAN~\cite{scan} uses splits of different sentence lengths, CFQ~\cite{CFQ} uses splits of different text structures, PCFG~\cite{PCFG} uses different split definitions to investigate different aspects of generalization, COGS~\cite{COGS} uses splits that can be addressed with compositional generalization, and the Mathematics dataset~\cite{math_dataset} uses extrapolation sets to measure generalization. \citet{pretrainedtransformersimproveOOD} showed that pretrained transformers help OOD generalization compared to other language models. They use three sentiment analysis datasets, i.e., generalization between SST-2~\cite{SST-2} and IMDb~\cite{IMDb}, the Yelp Review dataset with food types as domains, the Amazon Review dataset~\cite{amazon_rev1,amazon_rev2} with domains composed of clothing categories. They also used three reading comprehension datasets, i.e., STS-B~\cite{STS-B} has text of different genres (news and captions), ReCoRD~\cite{record} has news paragraphs from different news sources (CNN and Daily Mail), and MNLI~\cite{MNLI} has text from differently communicated interactions such as transcribed telephone and face-to-face conversations.

\paragraph{Temporal datasets} Some works looked at temporal distribution shifts in different settings. In natural language processing, \citet{mindthegap} investigated the ability of language models to generalize to future utterances beyond their training period on the WMT~\cite{wmt} and ArXiv~\cite{openarxiv} datasets. In the clinical setting, both \citet{empiricalframeworkclinical} and \citet{dg-temporaldataset-shift} investigated shifts when data is grouped according to the year in which they were gathered: the former used in-hospital mortality records and X-rays of the lungs, while the later used patients health record in the ICU. \citet{shifts} investigated temporal shifts in large amounts of weather data.

\paragraph{Other modalities} There have been efforts in studying OOD generalization on graphs, such as works from \citet{ood-on-graph} and the OGB-MolPCBA~\cite{wilds} dataset adapted from the Open Graph Benchmark~\cite{opengraph}.

As mentionned in Section~\ref{sect:intro}, multiple works focused on gathering and standardizing datasets for a unified measure of OOD generalization algorithm performance. \citet{domainbed} introduced DomainBed: a collection of seven image datasets (i.e., CMNIST, RMNIST, PACS, VLCS, Office-Home, Terra Incognita, DomainNet) for a systematic OOD performance evaluation of algorithms. \citet{ood-bench} built on top of DomainBed and added three datasets (i.e., Camelyon17-WILDS, NICO, and CelebA), along with a measure to group the datasets according to their distribution shift. \citet{wilds} introduced WILDS: a benchmark of several new in-the-wild distribution shifts datasets across diverse data modalities, i.e., IWildCam2020-WILDS, Camelyon17-WILDS, RxRx1-WILDS, OGB-MolPCBA, GlobalWheat-WILDS, CivilComments-WILDS, FMoW-WILDS, PovertyMap-WILDS, Amazon-WILDS, and Py150-WILDS. WILDS was recently extended with unlabeled samples for multiple of its datasets~\cite{WILDS-v2}.

\section{Additional dataset information} \label{appendix:datasets}

\subsection{Spurious-Fourier} \label{appendix:spurious_fourier}

\begin{figure}[H]
    \centering
    \includegraphics[width=\linewidth]{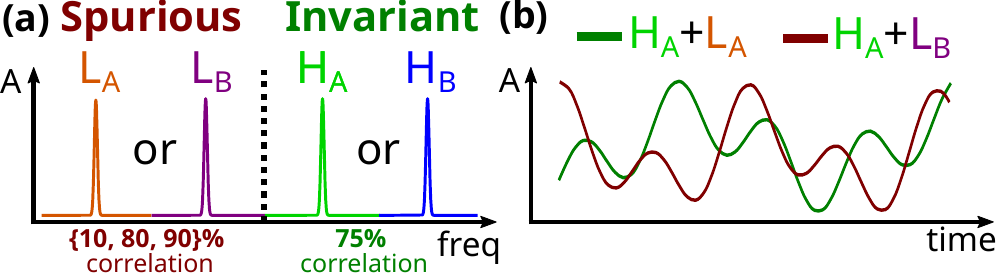}
    \caption{Description of the Spurious-Fourier dataset. Signals have one low-frequency peak and one high-frequency peak. They are then constructed from the Fourier spectrum with an inverse Fourier transform. (b) Examples of reconstructed signals, both signals have the same high frequency, but different low frequencies, which are hard to distinguish visually.}
    \label{fig:spurious_fourier_appendix}
\end{figure}

\subsubsection{Setup}

\paragraph{Motivation}
Recall the cow or camel classification problem from Section~\ref{sect:intro}, where a deep learning model trained to distinguish cows from camels learns to rely on the background properties (e.g., grass or sand) instead of the animal characteristic features (e.g., color) to make a prediction. \citet{IRM} proposed Colored MNIST (CMNIST) to recreate the the cow or camel classification problem into a simple benchmark in the image domain. We propose the Spurious-Fourier dataset which is an adaptation of the cow or camel classification problem to time series.

\paragraph{Problem setting} 
We create a dataset composed of one-dimensional signals, where the task is to perform binary classification based on the frequency characteristics. Signals are constructed from Fourier spectra with one low-frequency peak ($L_A=2$Hz or $L_B=4$Hz) and one high-frequency peak ($H_A=7$Hz or $H_B=9$Hz), see Figure \ref{fig:spurious_fourier_appendix}. Domains $D^d|_{d\in\{10\%,\,80\%,\,90\%\}}$ contain signal-label pairs, where the label is a noisy function of the low- and high-frequencies such that low-frequency peaks bear a varying correlation of $d$ with the label and high-frequency peaks bear an invariant correlation of $75\%$ with the label.

\paragraph{Data}
We first create four Fourier spectra with all combinations of low- and high-frequency peaks. From each of the spectra, we perform an inverse Fourier transform to get a 1 dimensional signal of 100 seconds sampled at 100Hz. We then split this long signal into smaller overlapping sequences of 50 time-steps, i.e., half a second. We then recreate the Colored MNIST~\cite{IRM} dataset characteristic. We build datasets $D^d$ by repeating the following protocol 4000 times. First, we sample $y$ from a Bernoulli distribution $p=0.5$. Second, we obtain $\tilde y$ by flipping $y$ with a probability of $25\%$, this gives us our high-frequency component $h$ ($\tilde y=0 \rightarrow H_A$, $\tilde y=1 \rightarrow H_B$). Third, we sample $z$ from a Bernoulli distribution of parameter $p=d$, this gives us our low-frequency component $l$ ($z=0 \rightarrow L_A$, $z=1 \rightarrow L_B$). Finally, we add to the domains dataset $D^d$ a random signal of configuration $l+h$ with the label $\tilde y$.

\paragraph{Domain information} 
Table~\ref{table:spurious_fourier_domain_proportions} details the distribution of labels for every domain in the Spurious-Fourier dataset.

\begin{table}[H]
    \centering
    \caption{Distribution of labels for every domain in the Spurious-Fourier dataset}
    \begin{tabular}{lccc}
        \toprule
        \textbf{Domain} & 7Hz & 9Hz & \textbf{Total} \\
        \midrule
        10\% &  2043 & 1957 & 4000 \\
        80\% &  2013 & 1987 & 4000 \\
        90\% &  1991 & 2009 & 4000 \\
        \midrule
        \textbf{Total} & 6047 & 5953 & 12000\\
        \bottomrule
    \end{tabular}
    \label{table:spurious_fourier_domain_proportions}
\end{table}

\paragraph{Architecture choice}

For this simple task, we use the LSTM~\cite{lstm} model because it is a simple model well accepted in the time series/sequential prediction field. We stack on top of the LSTM a fully connected (FC) layer used to make predictions at the last time step of the time series. Layers are detailed in Table \ref{table:SF_model}

\begin{table}[H]
    \centering
    \caption{Model architecture used for the Spurious-Fourier dataset}
    \begin{tabular}{cl}
        \toprule
        \textbf{\#} & Layer \\
        \midrule
        15 & LSTM(in=1, hidden\_size=20, num\_layers=2)\\
        16 & Linear(in=20, out=20)\\
        17 & ReLU\\
        18 & Linear(in=20, out=2)\\
        \bottomrule
    \end{tabular}
    \label{table:SF_model}
\end{table}

\subsubsection{Detailed results}
\paragraph{Oracle task}
Investigating the impact of spurious correlation in a dataset is meaningless if the underlying invariant task is impossible to solve with a given model or hyperparameter configuration. In order to avoid this, we provide the Basic-Fourier dataset in the WOODS repository. It consists of the oracle task of the Spurious-Fourier dataset, i.e., classifying $7$Hz and $9$Hz signals with no label noise or spurious features. We create two Fourier spectra with $7$Hz and $9$Hz frequency peaks respectively. From both of the spectra, we perform an inverse Fourier transform to get a one-dimensional signal of 100 seconds sampled at 100Hz. We then split this long signal into smaller overlapping sequences of 50 time-steps, i.e., half a second. While this is not a domain generalization task, the Basic-Fourier dataset is included in the WOODS repository as a sanity check that the underlying invariant task of the Spurious-Fourier dataset is possible with the model and hyperparameter configuration we are using. We show the results of ERM on the Basic-Fourier dataset in Table~\ref{table:basic_fourier_results}.

\begin{table}[H]
    \centering
    \caption{Result for the Basic-Fourier dataset.}
    \adjustbox{max width=\textwidth}{%
        \begin{tabular}{lcc}
            \toprule
            \textbf{Objective} & Performance\\
            \midrule
            ERM &  $100.00 \, (0.00)$ \\
            \bottomrule
        \end{tabular}
    }
    \label{table:basic_fourier_results}
\end{table}

\paragraph{ID evaluation}
We evaluate the performance of ERM with access to all domains $D^d|_{d\in\{10\%,80\%,90\%\}}$. We obtain these results by doing a hyperparameter search with the methodology detailed in Appendix~\ref{appendix:framework} with no held-out test domain and choose the model with train-domain validation. In other words, the training is done with all domains; thus, all domains are ID. The columns correspond to the validation accuracy of the chosen model in each domain. We see that the model learns the invariant solution to the task because the high frequencies ($75\%$) are a stronger predictor of the label than the low frequencies ($60\%$).

\begin{table}[H]
    \centering
    \caption{ID results for wthe Spurious-Fourier dataset}
    \adjustbox{max width=\textwidth}{%
        \begin{tabular}{lccccc}
            \toprule
            \textbf{Algorithm} & 10\% & 80\% & 90\% & \textbf{Average}\\
            \midrule
            ID ERM &  $74.46 \, (0.07)$  &  $74.79 \, (0.03)$  &  $73.54 \, (0.07)$  &  74.26 \\
            \bottomrule
        \end{tabular}
    }
    \label{table:spurious_fourier_ID}
\end{table}

\paragraph{Benchmark results}
We present the detailed evaluation of OOD generalization algorithms on the Spurious-Fourier dataset. Important note: We evaluate performance only when holding out the $10\%$ domain as it is the only domain of meaning, and including the other domains only dilutes the information carried by this dataset.

\begin{table*}[h]
	\centering
	\caption{OOD generalization algorithms performance on the Spurious-Fourier dataset}
    \begin{minipage}{0.49\linewidth}
		\begin{center}
		\adjustbox{max width=\linewidth}{%
			\begin{tabular}{lc}
				\toprule
				\multicolumn{2}{c}{\textbf{Train-domain validation}} \\
				\midrule
				\textbf{Objective} & $10\%$ \\
				\midrule
				ERM &  $9.91 \, (0.12)$ \\
				IRM &  $10.30 \, (0.09)$  \\
				VREx &  $10.36 \, (0.23)$ \\
				GroupDRO &  $10.06 \, (0.19)$  \\
				IB-ERM &  $9.21 \, (0.31)$  \\
				SD &  $9.67 \, (0.20)$ \\
				\bottomrule
			\end{tabular}}
		\end{center}
	\end{minipage}
	\begin{minipage}{0.49\linewidth}
		\begin{center}
		\adjustbox{max width=\linewidth}{%
			\begin{tabular}{lc}
				\toprule
				\multicolumn{2}{c}{\textbf{Test-domain validation}} \\
				\midrule
				\textbf{Objective} & $10\%$ \\
				\midrule
				ERM &  $12.07 \, (1.99)$  \\
				IRM &  $58.82 \, (1.98)$  \\
				VREx &  $63.69 \, (0.70)$  \\
				GroupDRO &  $21.49 \, (2.12)$  \\
				IB-ERM &  $18.65 \, (4.01)$  \\
				SD &  $9.97 \, (0.11)$  \\
				\bottomrule
			\end{tabular}}
		\end{center}
	\end{minipage}
	\label{table:detailed_spurious_fourier_results}
\end{table*}

\subsection{Temporal Colored MNIST with source domains} \label{appendix:tcmnist}

\begin{figure}[H]
    \centering
    \includegraphics[width=\linewidth]{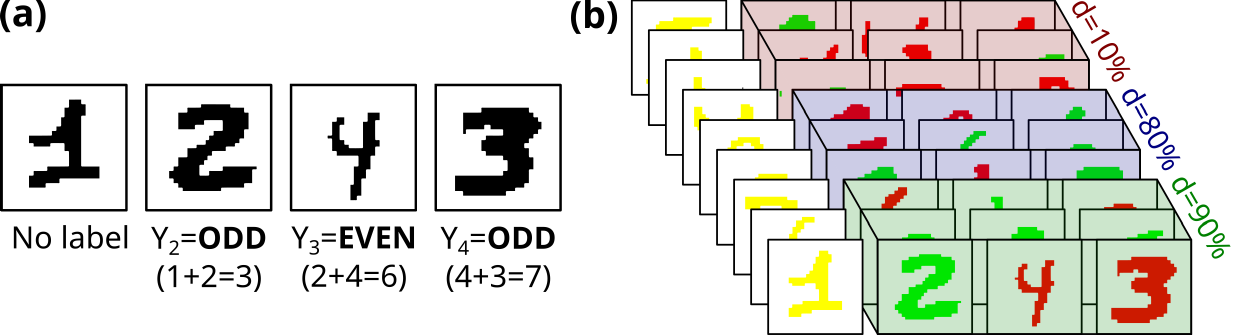}
    \caption{Description of the Temporal Colored MNIST dataset with source domains. (a) Data samples are videos of four colored MNIST digits where the task is to predict whether the sum of the current and previous digits in the sequence is odd or even. (b) Spuriously correlated color is added to each digit such that the correlation is constant among the frames of a video, but varies between video from different domains $d\in\{10\%,\,80\%,\,90\%\}$.}
    \label{fig:tcmnist_source_appendix}
\end{figure}

\subsubsection{Setup}

\paragraph{Motivation}
\citet{IRM} proposed the CMNIST dataset as a synthetic investigation of the cow or camel classification problem. We propose an extension of this widely used dataset to time series to investigate both domain definition paradigms presented in Section~\ref{sect:problem_formulation_time_series}: Source-domains (Example~\ref{ex:source_domains}) and Time-domains (Example~\ref{ex:time_domains}). In this section, we give more details on the Source-domain formulation of the the dataset.

\paragraph{Problem setting}
In Temporal Colored MNIST with source domains (TCMNIST-Source), we create a binary classification task of video frames. Videos are sequences of four colored MNIST digits where the goal is to predict whether the sum of the current and previous digits in the sequence is odd or even, see Figure~\ref{fig:tcmnist_source_appendix}(a). Prediction is made for all frames except for the first one. The label is a noisy function of the digit and color, such that the color bears a varying correlation of $d$ with the label of the frame, and the digit sums bears an invariant correlation of $75\%$ with the label of the frame. Domains are created such that the color correlation is constant among the frames of a video, but varies between video from different domains $d\in\{10\%,\,80\%,\,90\%\}$. The domain definition is depicted in Figure~\ref{fig:tcmnist_source_appendix}(b).

\paragraph{Data}
We create videos by concatenating four digits together and attributing labels $\mathbf y$ to the second, third and fourth frames following the parity task, see Figure~\ref{fig:tcmnist_source_appendix}(a). For every labeled frame $i$ in a sequence from the domain $d\in\{10\%,\,80\%,\,90\%\}$, we define the final label of that frame $\tilde{\mathbf y}_i$ by flipping the label $\mathbf y_i$ with a probability of $25\%$. Second, we define $z$ as $\tilde{\mathbf y}_i$ flipped with a probability equals to the domain definition ($10\%$, $80\%$ or $90\%$). Finally, we color the digit red if $z=0$ or green if $z=1$.

\paragraph{Domain information} 
Table~\ref{table:tcmnist_source_domain_proportions} details the distribution of labels for every domain in the TCMNIST-Source dataset.
\begin{table}[H]
    \centering
    \caption{Distribution of labels for every domain in the TCMNIST-Source dataset}
    \centering
    \begin{tabular}{lccc}
        \toprule
        \textbf{Domain} & Even & Odd & \textbf{Domain Total} \\
        \midrule
        10\% &  8603 & 8899 & 17502 \\
        80\% &  8583 & 8916 & 17499 \\
        90\% &  8563 & 8936 & 17499 \\
        \midrule
        \textbf{Total} & 25749 & 26751 & 52500\\
        \bottomrule
    \end{tabular}
    \label{table:tcmnist_source_domain_proportions}
\end{table}

\paragraph{Architecture choice}
For this task, we use a combination of a CNN and an LSTM architecture. Table~\ref{table:TCMNIST_arch} details the layers of the model architecture. Its parameters were hand tuned to perform well on this task.

\begin{table}[H] 
    \centering
    \caption{Model architecture used for the TCMNIST-Source dataset}
    \begin{tabular}{cl}
        \toprule
        \textbf{\#} & Layer \\
        \midrule
        1 & Conv2D(in=d, out=8, padding=1)\\
        2 & ReLU\\
        3 & Conv2D(in=8, out=32, stride=2, padding=1)\\
        4 & ReLU\\
        5 & MaxPool2d\\
        6 & Conv2D(in=32, out=32, padding=1)\\
        7 & ReLU\\
        8 & MaxPool2d\\
        9 & Conv2D(in=32, out=32, padding=1)\\
        10 & ReLU\\
        11 & Linear(in=288, out=64)\\
        12 & ReLU\\
        13 & Linear(in=64, out=32)\\
        14 & ReLU\\
        15 & LSTM(in=32, hidden\_size=128, num\_layers=1)\\
        16 & Linear(in=128, out=64)\\
        17 & ReLU\\
        16 & Linear(in=64, out=64)\\
        17 & ReLU\\
        18 & Linear(in=64, out=2)\\
        \bottomrule
    \end{tabular}
    \label{table:tcmnist_source_architecture}
\end{table}

\subsubsection{Detailed results}

\paragraph{Oracle task}
Investigating the impact of spurious correlation in a dataset is meaningless if the underlying invariant task is impossible to solve with a given model or hyperparameter configuration. In order to avoid this, we provide the Temporal MNIST (TMNIST) dataset in the WOODS repository. It consists of the oracle task of the TCMNIST-Source dataset, i.e., classifying whether the sum of the current and last digit is odd or even without label noise and without spurious features. We create videos by concatenating four digits together and attributing labels $\mathbf y$ to the second, third and fourth frames following the parity task. While this is not a domain generalization task, the TMNIST dataset is included in the WOODS repository as a sanity check that the underlying invariant task of the Spurious-Fourier dataset is possible with the model and hyperparameter configuration we are using. We show the results of ERM on the TMNIST dataset in Table~\ref{table:tmnist_source_results}.

\begin{table}[H]
    \centering
    \caption{Result for the TMNIST dataset}
    \adjustbox{max width=\textwidth}{%
        \begin{tabular}{lcc}
            \toprule
            \textbf{Objective} & Performance\\
            \midrule
            ERM &  $98.77 \, (0.02)$ \\
            \bottomrule
        \end{tabular}
    }
    \label{table:tmnist_source_results}
\end{table}

\paragraph{ID evaluation}
We show the ID results of ERM for TCMNIST-Source in Table~\ref{table:tcmnist_source_ID}. We obtain these results by doing a hyperparameter search with the methodology detailed in Appendix~\ref{appendix:framework} with no held-out test domain and choose the model with train-domain validation. In other words, the training is done with all domains; thus, all domains are ID. The columns correspond to the validation accuracy of the chosen model in each domain. 

\begin{table}[H]
    \centering        
    \caption{ID results for the TCMNIST-Source dataset}
    \adjustbox{max width=\textwidth}{%
        \begin{tabular}{lccccc}
            \toprule
            \textbf{Algorithm} & 10\% & 80\% & 90\% & \textbf{Average}\\
            \midrule
            ID ERM &  $68.36 \, (0.13)$  &  $73.49 \, (0.13)$  &  $74.85 \, (0.16)$  &  72.23 \\
            \bottomrule
        \end{tabular}
    }
    \label{table:tcmnist_source_ID}
\end{table}

\paragraph{Benchmark results}
We show the detailed benchmark results of the adapted OOD generalization algorithms in Table~\ref{table:detailed_tcmnist_source_results}.
\begin{table*}[h]
	\centering
	\caption{OOD generalization algorithms performance on the TCMNIST-Source dataset}
	\begin{minipage}{0.49\linewidth}
		\begin{center}
		\adjustbox{max width=\linewidth}{%
			\begin{tabular}{lc}
				\toprule
				\multicolumn{2}{c}{\textbf{Train-domain validation}} \\
				\midrule
				\textbf{Objective} & 10\% \\
				\midrule
				ERM &  $10.07 \, (0.02)$ \\
				IRM &  $9.82 \, (0.13)$ \\
				VREx &  $9.83 \, (0.22)$ \\
				GroupDRO &  $10.39 \, (0.02)$ \\
				IB-ERM &  $9.97 \, (0.07)$ \\
				SD &  $10.24 \, (0.12)$ \\
				\bottomrule
			\end{tabular}}
		\end{center}
	\end{minipage}
	\centering
	\begin{minipage}{0.49\linewidth}
		\begin{center}
		\adjustbox{max width=\linewidth}{%
			\begin{tabular}{lc}
				\toprule
				\multicolumn{2}{c}{\textbf{Test-domain validation}} \\
				\midrule
				\textbf{Objective} & 10\% \\
				\midrule
				ERM &  $30.34 \, (0.82)$ \\
				IRM &  $52.74 \, (0.59)$ \\
				VREx &  $49.69 \, (0.25)$ \\
				GroupDRO &  $33.52 \, (2.95)$ \\
				IB-ERM &  $28.12 \, (1.12)$ \\
				SD &  $27.35 \, (3.51)$ \\
				\bottomrule
			\end{tabular}}
		\end{center}
	\end{minipage}
	\label{table:detailed_tcmnist_source_results}
\end{table*}

\subsection{Temporal colored MNIST with time domains}

\begin{figure}[H]
    \centering
    \includegraphics[width=\linewidth]{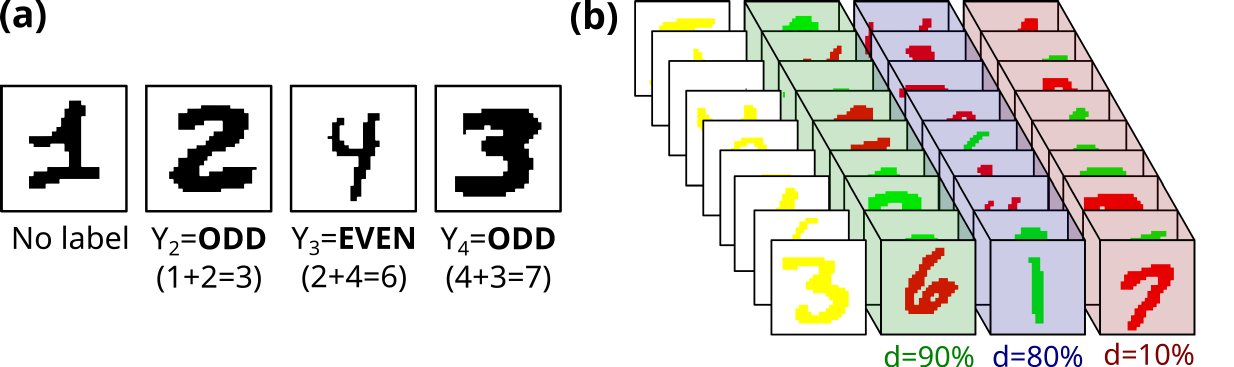}
    \caption{Description of the Temporal Colored MNIST dataset with time domains. (a) Data samples are videos of four colored MNIST digits where the task is to predict whether the sum of the current and previous digits in the sequence is odd or even. (b) Spuriously correlated color is added to each digit such that the correlation varies across frames. However, videos all have the same sequence of color correlation, where the first labeled frame correlation is $90\%$, second is $80\%$ and third is $10\%$.}
    \label{fig:tcmnist_time_appendix}
\end{figure}

\subsubsection{Setup}

\paragraph{Motivation}
\citet{IRM} proposed the CMNIST dataset as a synthetic investigation of the cow or camel classification problem. We propose an extension of this widely used dataset to time series to investigate both domain definition paradigms presented in Section~\ref{sect:problem_formulation_time_series}: Source-domains (Example~\ref{ex:source_domains}) and Time-domains (Example~\ref{ex:time_domains}). In this section, we give more details on the Time-domain formulation of the the dataset.

\paragraph{Problem setting}
In Temporal Colored MNIST with time domains (TCMNIST-Time), we create a binary classification task of video frames. Videos are sequences of four colored MNIST digits where the goal is to predict whether the sum of the current and previous digits in the sequence is odd or even, see Figure~\ref{fig:tcmnist_time_appendix}(a). Prediction is made for all frames except for the first one. The label is a noisy function of the digit and color, such that the color bears a varying correlation of $d$ with the label of the frame, and the digit sums bears an invariant correlation of $75\%$ with the label of the frame. Domains are created such that the color correlation varies across frames. However, videos all have the same sequence of color correlation, where the first labeled frame correlation is $90\%$, second is $80\%$ and third is $10\%$. The domain definition is depicted in Figure~\ref{fig:tcmnist_time_appendix}(b).

\paragraph{Data}
We create videos by concatenating four digits together and attributing labels $\mathbf y$ to the second, third and fourth frames following the parity task, see Figure~\ref{fig:tcmnist_time_appendix}(a). For every labeled frame $i\in \{2,3,4\}$ of all videos in the dataset, we define the final label of that frame $\tilde{\mathbf y}_i$ by flipping the label $\mathbf y_i$ with a probability of $25\%$. Second, we define $z$ as $\tilde{\mathbf y}_i$ flipped with a probability equals to the domain definition for that frame index ($i=2 \rightarrow 90\%$, $i=3 \rightarrow 80\%$ or $i=4 \rightarrow 10\%$). Finally, we color the digit red if $z=0$ or green if $z=1$.

\paragraph{Domain information}
Table~\ref{table:tcmnist_time_domain_proportions} details the distribution of labels for every domain in the TCMNIST-Time dataset.
\begin{table}[H]
    \centering
    \caption{Distribution of labels for every domain in the TCMNIST-Time dataset}
    \begin{minipage}{\linewidth}
        \centering
        \begin{tabular}{lccc}
            \toprule
            \textbf{Domain} & Even & Odd & \textbf{Domain Total} \\
            \midrule
            10\% &  8564 & 8936 & 17500 \\
            80\% &  8765 & 8735 & 17500 \\
            90\% &  8613 & 8887 & 17500 \\
            \midrule
            \textbf{Total} & 25942 & 26558 & 52500\\
            \bottomrule
        \end{tabular}
    \end{minipage}
    \label{table:tcmnist_time_domain_proportions}
\end{table}

\paragraph{Architecture choice}
For this task, we use a combination of a CNN and an LSTM architecture. Table~\ref{table:TCMNIST_arch} details the layers of the model architecture. Its parameters were hand tuned to perform well on this toy task.

\begin{table}[H] 
    \centering
    \caption{Model architecture used for the TCMNIST-Time dataset}
    \begin{tabular}{cl}
        \toprule
        \textbf{\#} & Layer \\
        \midrule
        1 & Conv2D(in=d, out=8, padding=1)\\
        2 & ReLU\\
        3 & Conv2D(in=8, out=32, stride=2, padding=1)\\
        4 & ReLU\\
        5 & MaxPool2d\\
        6 & Conv2D(in=32, out=32, padding=1)\\
        7 & ReLU\\
        8 & MaxPool2d\\
        9 & Conv2D(in=32, out=32, padding=1)\\
        10 & ReLU\\
        11 & Linear(in=288, out=64)\\
        12 & ReLU\\
        13 & Linear(in=64, out=32)\\
        14 & ReLU\\
        15 & LSTM(in=32, hidden\_size=128, num\_layers=1)\\
        16 & Linear(in=128, out=64)\\
        17 & ReLU\\
        16 & Linear(in=64, out=64)\\
        17 & ReLU\\
        18 & Linear(in=64, out=2)\\
        \bottomrule
    \end{tabular}
    \label{table:TCMNIST_arch}
\end{table}

\subsubsection{Detailed results}

\paragraph{Oracle task}
Investigating the impact of spurious correlation in a dataset is meaningless if the underlying invariant task is impossible to solve with a given model or hyperparameter configuration. In order to avoid this, we provide the Temporal MNIST (TMNIST) dataset in the WOODS repository. It consists of the oracle task of the TCMNIST-Time dataset, i.e., classifying whether the sum of the current and last digit is odd or even without label noise and without spurious features. We create videos by concatenating four digits together and attributing labels $\mathbf y$ to the second, third and fourth frames following the parity task. While this is not a domain generalization task, the TMNIST dataset is included in the WOODS repository as a sanity check that the underlying invariant task of the Spurious-Fourier dataset is possible with the model and hyperparameter configuration we are using. We show the results of ERM on the TMNIST dataset in Table~\ref{table:tmnist_time_results}.

\begin{table}[H]
    \centering
    \caption{Result for the TMNIST dataset}
    \adjustbox{max width=\textwidth}{%
        \begin{tabular}{lcc}
            \toprule
            \textbf{Objective} & Performance\\
            \midrule
            ERM &  $98.77 \, (0.02)$ \\
            \bottomrule
        \end{tabular}
    }
    \label{table:tmnist_time_results}
\end{table}

\paragraph{ID evaluation}
We show the ID results of ERM for TCMNIST-Time in Table~\ref{table:tcmnist_time_ID}. We obtain these results by doing a hyperparameter search with the methodology detailed in Appendix~\ref{appendix:framework} with no held-out test domain and choose the model with train-domain validation. In other words, the training is done with all domains; thus, all domains are ID. The columns correspond to the validation accuracy of the chosen model in each domain. 

\begin{table}[H]
    \centering
    \caption{ID results for the TCMNIST-Time dataset}
    \adjustbox{max width=\textwidth}{%
        \begin{tabular}{lccccc}
            \toprule
            \textbf{Algorithm} & 10\% & 80\% & 90\% & \textbf{Average}\\
            \midrule
            ID ERM &  $89.97 \, (0.00)$  &  $80.98 \, (0.02)$  &  $91.20 \, (0.00)$  &  87.38 \\
            \bottomrule
        \end{tabular}
    }
    \label{table:tcmnist_time_ID}
\end{table}

\paragraph{Benchmark results}
We show the detailed benchmark results of the adapted OOD generalization algorithms in Table~\ref{table:detailed_tcmnist_time_results}.

\begin{table*}[h]
	\centering
	\caption{OOD generalization algorithms performance on the TCMNIST-Time dataset}
	\begin{minipage}{0.49\linewidth}
		\begin{center}
		\adjustbox{max width=\linewidth}{%
			\begin{tabular}{lc}
				\toprule
				\multicolumn{2}{c}{\textbf{Train-domain validation}} \\
				\midrule
				\textbf{Objective} & 10\% \\
				\midrule
				ERM &  $9.88 \, (0.11)$  \\
				GroupDRO &  $10.09 \, (0.14)$  \\
				IB-ERM &  $10.19 \, (0.04)$ \\
				IRM &  $10.27 \, (0.05)$  \\
				SD &  $10.19 \, (0.14)$  \\
				VREx &  $9.80 \, (0.06)$ \\
				\bottomrule
			\end{tabular}}
		\end{center}
	\end{minipage}
	\centering
	\begin{minipage}{0.49\linewidth}
		\begin{center}
		\adjustbox{max width=\linewidth}{%
			\begin{tabular}{lc}
				\toprule
				\multicolumn{2}{c}{\textbf{Test-domain validation}} \\
				\midrule
				\textbf{Objective} & 10\% \\
				\midrule
				ERM &  $28.61 \, (2.41)$ \\
				GroupDRO &  $24.85 \, (3.91)$ \\
				IB-ERM &  $33.70 \, (6.49)$  \\
				IRM &  $50.65 \, (0.17)$  \\
				SD &  $31.76 \, (5.15)$ \\
				VREx &  $50.57 \, (0.59)$ \\
				\bottomrule
			\end{tabular}}
		\end{center}
	\end{minipage}
	\label{table:detailed_tcmnist_time_results}
\end{table*}

\subsection{CAP} \label{appendix:cap}

\begin{figure}[H]
    \centering
    \includegraphics[width=0.8\linewidth]{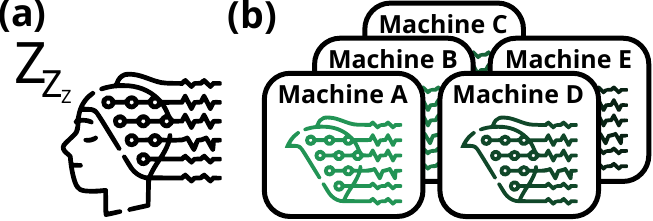}
    \caption{Summary of the CAP dataset. (a) The task is to perform sleep stage classification from EEG measurements. (b) The dataset has five source domains, where each domain contains data gathered with a different machine. The goal is to generalize to unseen machines.}
    \label{fig:cap_appendix}
\end{figure}

\subsubsection{Setup}

\paragraph{Motivation}
A recurrent problem in computational medicine is that models trained on data from a given recording device will not generalize to data coming from another device, even when both devices are from a similar equipment provider. Failure to generalize to unseen machines can cause critical issues for clinical practice because a false sense of confidence in a model could lead to a false diagnosis~\cite{repro_in_bioinformatics,robust-EEG}. We study these machinery-induced distribution shifts with the CAP~\cite{cap,physionet} dataset (Figure~\ref{fig:cap_main_body}).

\paragraph{Problem setting}
We consider the sleep stage classification task from electroencephalographic (EEG) measurements. The dataset has five source domains, where each domain contains data gathered with a different machine. The goal is to generalize to unseen machines.

\paragraph{Data} 
The dataset is composed of $40\,390$ gathered on 41 participants. Each participants had one night of sleep recorded. The inputs $\mathbf X$ are recordings of 30 seconds each with 19 channels sampled at 100Hz. The channels include EEG but also include Electromyography (EMG), Electrocardiography (ECG), and heart rate measurements. The labels $\mathbf Y$ consist of 6 sleep stages: Awake, Non-REM 1, Non-REM 2, Non-REM 3, Non-REM 4, and REM. The domains $d$ are the 5 EEG machines: Machine A, Machine B, Machine C, Machine D, and Machine E.

\paragraph{Preprocessing}
This section details the preprocessing steps taken for the CAP dataset. The raw CAP dataset contains data from 15 machines, each with different channels and sampling frequency characteristics. We only use recordings from the five machines with the most data. Removing machines with less data allows us to retain a reasonable number (19) of shared channels between them. Next, we resample the data to a standard sampling frequency of 100Hz for all five machines. We then apply a bandpass filter from 0.3Hz to 30Hz. This bandpass filter removes frequency bands generally considered uninformative for sleep stage classification. Next, we split the nights of sleep into sequences of 30 seconds for training and testing. Finally, we then detrend and normalize the 30 second recordings with a standard scaler applied to the channels individually.

\paragraph{Domain information}
Table~\ref{table:cap_domain_participants} details the number of participants per domain and some demographic information; each had a single night of sleep recorded. Table~\ref{table:cap_domain_proportions} details the proportion of samples and labels across domains.

\begin{table}[H]
    \centering
    \caption{Number of participants and demographic information of the CAP dataset}
    \begin{tabular}{lc|cc|c}
        \toprule
        \textbf{Domain} & Number of participants & Male & Female & Age  \\
        \midrule
        Machine A      & 13 & 7 & 6 & $33.1\pm 13.9$ \\
        Machine B      & 5 & 5 & 0 & $26.4\pm 8.2$ \\
        Machine C      & 5 & 4 & 1 & $73.4\pm 6.42$ \\
        Machine D      & 10 & 5 & 5 & $30.7 \, (8.9$ \\
        Machine E      & 8 & 3 & 5 & $36.8\pm 16.7$ \\
        \midrule
        \textbf{Total} & 41 & 24 & 17 & $37.3\pm 18.1$ \\
        \bottomrule
    \end{tabular}
    \label{table:cap_domain_participants}
\end{table}
\begin{table}[H]
    \centering
    \caption{Domain proportions of labels in the CAP dataset}
    \begin{tabular}{lccccccc}
        \toprule
        \textbf{Domain} & Awake & NREM 1 & NREM 2 & NREM 3 & NREM 4 & REM & \textbf{Domain Total} \\
        \midrule
        Machine A      & 1448 & 350  &  4986    &  1533 &  2110 & 2342  & 12769 \\
        Machine B      & 318  & 171  &  1933    &  595  &  706  & 971   & 4694 \\
        Machine C      & 1318 & 294  &  1168    &  595  &  810  & 547   & 4732 \\
        Machine D      & 1114 & 580  &  3547    &  1273 &  1606 & 1810  & 9930 \\
        Machine E      & 967  & 276  &  3377    &  711  &  1251 & 1683  & 8265 \\
        \midrule
        \textbf{Total} & 5165 & 1671 &  15011   &  4707 &  6483 & 7353  & 40390\\
        \bottomrule
    \end{tabular}
    \label{table:cap_domain_proportions}
\end{table}

\paragraph{Architecture choice}
For this dataset, we use a deep convolution network model as defined in work from~\citet{braindecode}. We use the implementation of the BrainDecode~\cite{braindecode} Toolbox. We chose this model because it is the perfect combination of performance stability and recognition from the EEG community. The implementation is available at \url{https://github.com/TNTLFreiburg/braindecode}.

\subsubsection{Detailed results}
\paragraph{ID evaluation}
We show the results of ERM for the CAP dataset in Table~\ref{table:cap_ID}. We obtain these results by doing a hyperparameter search with the methodology detailed in Appendix~\ref{appendix:framework} with no held-out test domain and choose the model with train-domain validation. In other words, the training is done with all domains; thus, all domains are ID. The columns correspond to the validation accuracy of the chosen model in each domain. 

\begin{table}[H]
    \centering
    \caption{ID results for the CAP dataset}
    \adjustbox{max width=\textwidth}{%
        \begin{tabular}{lccccccc}
            \toprule
            \textbf{Algorithm} & Machine A & Machine B & Machine C & Machine D & Machine E & \textbf{Average}\\
            \midrule
            ID ERM &  $78.26 \, (0.52)$  &  $78.14 \, (1.00)$  &  $63.39 \, (1.38)$  &  $78.73 \, (0.34)$  &  $77.09 \, (0.37)$  &  75.12 \\
            \bottomrule
        \end{tabular}
    }
    \label{table:cap_ID}
\end{table}

\paragraph{Benchmark results}
We show the detailed benchmark results of the adapted OOD generalization algorithms in Table~\ref{table:detailed_cap_results}. Each results is obtained by holding out one domain during training and reporting the performance of the chosen model from the hyperparameter sweep on that held out domain, more details in Appendix~\ref{appendix:framework}.

\begin{table*}[h]
	\centering
	\caption{OOD generalization algorithms performance on the CAP dataset}
	\begin{minipage}{\linewidth}
		\begin{center}
		\adjustbox{max width=\textwidth}{%
			\begin{tabular}{lcccccc}
				\toprule
				\multicolumn{7}{c}{\textbf{Train-domain validation}} \\
				\midrule
				\textbf{Objective} & Machine A & Machine B & Machine C & Machine D & Machine E & \textbf{Average}\\
				\midrule
				ERM &  $68.93 \, (0.54)$  &  $61.98 \, (0.53)$  &  $40.10 \, (0.75)$  &  $73.10 \, (0.83)$  &  $70.13 \, (0.33)$  &  62.85 \\
				IRM &  $67.59 \, (0.55)$  &  $48.40 \, (3.14)$  &  $41.01 \, (1.10)$  &  $69.52 \, (1.03)$  &  $66.86 \, (0.71)$  &  58.68 \\
				VREx &  $57.97 \, (1.92)$  &  $38.96 \, (0.42)$  &  $33.81 \, (1.19)$  &  $52.53 \, (3.49)$  &  $59.71 \, (1.53)$  &  48.60 \\
				GroupDRO &  $68.07 \, (0.33)$  &  $59.22 \, (1.53)$  &  $41.38 \, (0.52)$  &  $72.25 \, (0.70)$  &  $69.12 \, (0.90)$  &  62.01 \\
				IB-ERM &  $70.20 \, (0.71)$  &  $62.03 \, (1.79)$  &  $40.66 \, (0.58)$  &  $72.73 \, (0.18)$  &  $70.57 \, (0.83)$  &  63.24 \\
                SD &  $69.29 \, (0.25)$  &  $55.53 \, (1.45)$  &  $41.36 \, (1.78)$  &  $71.14 \, (0.22)$  &  $66.48 \, (0.92)$  &  60.76 \\
				\bottomrule
			\end{tabular}}
		\end{center}
	\end{minipage}
	\centering
	\begin{minipage}{\linewidth}
		\begin{center}
		\adjustbox{max width=\textwidth}{%
			\begin{tabular}{lcccccc}
				\toprule
				\multicolumn{7}{c}{\textbf{Oracle train-domain validation}} \\
				\midrule
				\textbf{Objective} & Machine A & Machine B & Machine C & Machine D & Machine E & \textbf{Average}\\
				\midrule
				ERM &  $69.00 \, (0.51)$  &  $65.21 \, (1.38)$  &  $43.11 \, (0.30)$  &  $73.31 \, (0.67)$  &  $70.34 \, (0.16)$  &  64.19 \\
				IRM &  $67.59 \, (0.55)$  &  $55.09 \, (2.08)$  &  $41.20 \, (1.19)$  &  $70.72 \, (0.46)$  &  $67.87 \, (0.26)$  &  60.49 \\
				VREx &  $57.79 \, (2.06)$  &  $39.49 \, (0.65)$  &  $36.38 \, (0.57)$  &  $52.95 \, (3.15)$  &  $59.71 \, (1.53)$  &  49.26 \\
				GroupDRO &  $68.73 \, (0.22)$  &  $60.39 \, (1.23)$  &  $43.19 \, (0.56)$  &  $72.51 \, (0.49)$  &  $69.91 \, (0.50)$  &  62.95 \\
				IB-ERM &  $71.06 \, (0.37)$  &  $66.99 \, (0.93)$  &  $43.21 \, (0.76)$  &  $73.64 \, (0.50)$  &  $70.88 \, (0.59)$  &  65.16 \\
                SD &  $69.38 \, (0.20)$  &  $61.84 \, (0.39)$  &  $43.97 \, (1.09)$  &  $71.41 \, (0.14)$  &  $69.41 \, (0.29)$  &  63.20 \\
				\bottomrule
			\end{tabular}}
		\end{center}
	\end{minipage}
	\label{table:detailed_cap_results}
\end{table*}

\subsubsection{Credit and license}

This dataset is adapted from the work of~\citet{cap}, as made available on the online Physionet~\cite{physionet} platform.
This dataset is licensed under the Open Data Commons Attribution License v1.0.

\subsection{SEDFx} \label{appendix:sedfx}

\begin{figure}[H]
    \centering
    \includegraphics[width=0.8\linewidth]{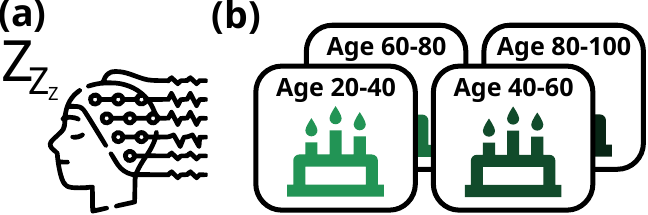}
    \caption{Summary of the SEDFx dataset. (a) The task is to perform sleep stage classification from EEG measurements. (b) The dataset has four source domains, where each domain contains data from participants of a certain age group. The goal is to generalize to unseen age groups.}
    \label{fig:sedfx_appendix}
\end{figure}

\subsubsection{Setup}

\paragraph{Motivation}
In clinical settings, we train a model on the data gathered from a limited number of patients and hope this model will generalize to new patients in the future~\cite{worstgroup_subpopulation}. However, this generalization between observed patients in the training dataset and new patients is not guaranteed. Distribution shifts caused by shifts in patient demographics (e.g., age, gender, and ethnicity) can cause the model to fail. We study age demographic shift with the SEDFx~\cite{sedfx,physionet} dataset (Figure~\ref{fig:sedfx_appendix}). 

\paragraph{Problem setting}
We consider the sleep classification task from EEG measurements. The dataset has four source domains, where each domain contains data from participants of a certain age group. The goal is to generalize to an unseen age demographic.

\paragraph{Data}
The dataset is composed of $238\,712$ recordings gathered on 100 participants. Every participant had 2 nights of sleep recorded. The inputs $\mathbf X$ are recordings of 30 seconds each with four EEG channels sampled at 100Hz. The channels include 2 EEG channels, one Electromyography (EOG) channel, and one Electrocardiography (ECG) channel. The labels $\mathbf Y$ consist of 6 sleep stages: Awake, Non-REM 1, Non-REM 2, Non-REM 3, Non-REM 4, and REM. The domains $d$ are the four disjoint age groups: Age 20-40, Age 40-60, Age 60-80, and age 80-100.

\paragraph{Preprocessing}
This section details the preprocessing steps taken for the SEDFx dataset. The raw SEDFx dataset contains data from 2 machines with different channels and sampling frequency characteristics. We use the data from both machines and keep only the four channels they have in common. First, we resample the data to a standard sampling frequency of 100Hz for both machines. We then apply a bandpass filter from 0.3Hz to 30Hz. This bandpass filter removes frequency bands generally considered uninformative for sleep stage classification. Next, we crop the unlabeled onset and end of the complete recordings. Next, we split the nights of sleep into shorter sequences of 30 seconds for training and testing. Finally, we detrend the data and normalize the 30 second recordings with a standard scaler applied to channels individually.

\paragraph{Domain information}
The data from the different machines consists of data from disjoint sets of participants. Table~\ref{table:sedfx_domain_proportions} details the number of participants per domain and some demographic information. Table~\ref{table:sedfx_domain_proportions} details the proportion of samples and labels across domains.

\begin{table}[H]
    \centering
    \caption{Number of participants and demographic information of the SEDFx dataset}
    \begin{tabular}{lc|cc|c}
        \toprule
        \textbf{Domain} & Number of participants & Male & Female & Age  \\
        \midrule
        Age 20-40   & 32 & 14 & 18 & $27.6\pm 4.7$ \\
        Age 40-60   & 29 & 12 & 17 & $53.3\pm 3.4$ \\
        Age 60-80   & 23 & 10 & 13 & $69.2\pm 3.5$ \\
        Age 80-100  & 16 & 8  & 8 & $90.5 \, (4.5$ \\
        \midrule
        \textbf{Total} & 100 & 44 & 56 & $54.7\pm 22.6$ \\
        \bottomrule
    \end{tabular}
    \label{table:sedfx_domain_participants}
\end{table}

\begin{table}[H]
    \centering
    \caption{Domain proportions of labels in the SEDFx dataset}
    \begin{tabular}{lccccccc}
        \toprule
        \textbf{Domain} & Awake & NREM 1 & NREM 2 & NREM 3 & NREM 4 & REM & \textbf{Domain Total} \\
        \midrule
        Age 20-40   & 10505  & 4222  &  28105    &  4830  &  4254  & 12348   & 64264 \\
        Age 40-60   & 20405  & 7182  &  27222    &  3243  &  1423  & 10007   & 69482 \\
        Age 60-80   & 14708  & 7087  &  19186    &  2830  &  1400  & 6917    & 52128 \\
        Age 80-100  & 25358  & 6684  &  14410    &  1288  &  186   & 4912    & 52838 \\
        \midrule
        \textbf{Total} & 70976  & 25175 &  88923    &  12191 &  7263  & 34184   & 238712\\
        \bottomrule
    \end{tabular}
    \label{table:sedfx_domain_proportions}
\end{table}

\paragraph{Architecture choice}
For this dataset, we use a deep convolution network model as defined in work from~\citet{braindecode}. We use the implementation of the BrainDecode~\cite{braindecode} Toolbox. We chose this model because it is the perfect combination of performance, stability, and recognition from the EEG community. The implementation is available at \url{https://github.com/TNTLFreiburg/braindecode}.

\subsubsection{Detailed results}

\paragraph{ID evaluation}
We show the results of ERM for the SEDFx dataset in Table~\ref{table:sedfx_ID}. We obtain these results by doing a hyperparameter search with the methodology detailed in Appendix~\ref{appendix:framework} with no held-out test domain and choose the model with train-domain validation. In other words, the training is done with all domains; thus, all domains are ID. The columns correspond to the validation accuracy of the chosen model in each domain. 

\begin{table}[H]
    \centering
    \caption{ID results for the SEDFx dataset}
    \adjustbox{max width=\textwidth}{%
        \begin{tabular}{lcccccc}
            \toprule
            \textbf{Algorithm} & Age 20-40 & Age 40-60 & Age 60-80 & Age 80-100 & \textbf{Average}\\
            \midrule
            ID ERM &  $74.11 \, (0.12)$  &  $74.19 \, (0.47)$  &  $72.39 \, (0.41)$  &  $69.22 \, (0.49)$  &  72.48 \\
            \bottomrule
        \end{tabular}
    }
    \label{table:sedfx_ID}
\end{table}

\paragraph{Benchmark results}
We show the detailed benchmark results of the adapted OOD generalization algorithms in Table~\ref{table:detailed_sedfx_results}. Each results is obtained by holding out one domain during training and reporting the performance of the chosen model from the hyperparameter sweep on that held out domain, more details in Appendix~\ref{appendix:framework}.

\begin{table*}[h]
	\centering
	\caption{OOD generalization algorithms performance on the SEDFx dataset}
	\begin{minipage}{\linewidth}
		\begin{center}
		\adjustbox{max width=\textwidth}{%
			\begin{tabular}{lccccc}
				\toprule
				\multicolumn{6}{c}{\textbf{Train-domain validation}} \\
				\midrule
				\textbf{Objective} & Age 20-40 & Age 40-60 & Age 60-80 & Age 80-100 & \textbf{Average}\\
				\midrule
				ERM &  $65.90 \, (2.06)$  &  $70.59 \, (0.54)$  &  $68.48 \, (0.17)$  &  $64.18 \, (0.33)$  &  67.29 \\
				IRM &  $60.76 \, (0.94)$  &  $67.69 \, (0.80)$  &  $63.04 \, (0.81)$  &  $59.13 \, (0.43)$  &  62.65 \\
				VREx &  $57.21 \, (2.39)$  &  $58.80 \, (0.94)$  &  $56.81 \, (0.97)$  &  $51.74 \, (1.23)$  &  56.14 \\
				GroupDRO &  $67.01 \, (0.90)$  &  $68.31 \, (0.73)$  &  $65.08 \, (0.42)$  &  $60.48 \, (1.07)$  &  65.22 \\
                IB-ERM &  $69.41 \, (0.12)$  &  $72.58 \, (0.46)$  &  $69.79 \, (0.42)$  &  $66.16 \, (0.87)$  &  69.48 \\
                SD &  $69.87 \, (1.34)$  &  $73.18 \, (0.19)$  &  $69.14 \, (0.16)$  &  $67.18 \, (0.26)$  &  69.84 \\
				\bottomrule
			\end{tabular}}
		\end{center}
	\end{minipage}
	\begin{minipage}{\linewidth}
		\begin{center}
		\adjustbox{max width=\textwidth}{%
			\begin{tabular}{lccccc}
				\toprule
				\multicolumn{6}{c}{\textbf{Oracle train-domain validation}} \\
				\midrule
				\textbf{Objective} & Age 20-40 & Age 40-60 & Age 60-80 & Age 80-100 & \textbf{Average}\\
				\midrule
				ERM &  $69.87 \, (0.41)$  &  $71.03 \, (0.20)$  &  $68.88 \, (0.18)$  &  $64.36 \, (0.26)$  &  68.53 \\
				IRM &  $66.02 \, (0.55)$  &  $67.69 \, (0.80)$  &  $63.16 \, (0.65)$  &  $60.14 \, (0.32)$  &  64.25 \\
				VREx &  $59.63 \, (0.76)$  &  $58.80 \, (0.94)$  &  $56.78 \, (0.95)$  &  $52.73 \, (0.22)$  &  56.99 \\
				GroupDRO &  $68.73 \, (0.37)$  &  $69.14 \, (0.40)$  &  $65.17 \, (0.35)$  &  $61.54 \, (0.85)$  &  66.15 \\
                IB-ERM &  $70.42 \, (0.41)$  &  $72.79 \, (0.62)$  &  $70.25 \, (0.12)$  &  $69.08 \, (0.56)$  &  70.64 \\
                SD &  $71.22 \, (0.60)$  &  $73.18 \, (0.19)$  &  $69.60 \, (0.04)$  &  $68.41 \, (0.60)$  &  70.60 \\
				\bottomrule
			\end{tabular}}
		\end{center}
	\end{minipage}
	\label{table:detailed_sedfx_results}
\end{table*}

\subsubsection{Credit and license}
This dataset was adapted from the work of~\citet{sedfx}, as made available on the online Physionet~\cite{physionet} platform. This dataset is licensed under the Open Data Commons Attribution license v1.0.

\subsection{PCL} \label{appendix:pcl}

\begin{figure}[H]
    \centering
    \includegraphics[width=0.8\linewidth]{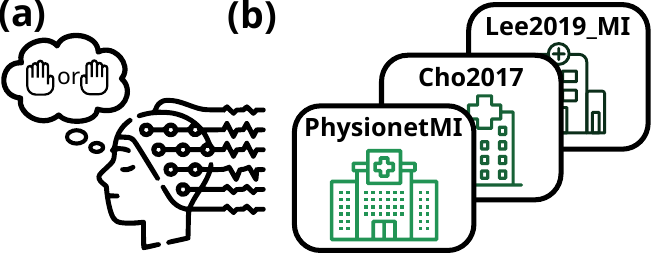}
    \caption{Summary of the PCL dataset. (a) The task is to perform motor imagery classification from EEG measurements. (b) The dataset has three source domains, where each domain contains a dataset from a different research group carrying out the same task. The goal is to generalize to unseen datasets of the same task.}
    \label{fig:pcl_appendix}
\end{figure}

\subsubsection{Setup}

\paragraph{Motivation}
Aside from changes in the recording device and shifts in patient demographics, human intervention in the data gathering process is another contributing factor to the distribution shift that can lead to failure of clinical models (e.g., Camelyon17~\cite{wilds,WILDS-v2}). This challenge is especially prevalent in temporal medical data (e.g., EEG, MEG, and others) because recording devices are complex tools greatly affected by nonlinear effects and modulations. These effects are often caused by context and preparations made before the recording~\cite{robust-EEG}. We study these procedural shifts with the PCL~\cite{lee2019,cho2017,physionetMI,moabb} dataset (Figure~\ref{fig:pcl_appendix}).

\paragraph{Problem setting}
We consider the motor imagery task from electroencephalographic (EEG) measurements. The dataset has three source domains, where each domain contains a dataset from a different research group carrying out the same task. The goal is to generalize to unseen data gathering processes.

\paragraph{Data}
The dataset is composed of $22\,598$ recordings gathered with 215 participants. The inputs $\mathbf X$ are recordings of three seconds each with 48 EEG channels sampled at 250Hz. The 48 channels contain only EEG measurements. The labels $\mathbf Y$ are two imagined movements: left hand and right hand. The domains $d$ are four different motor imagery datasets: Schalk04~\cite{physionetMI}, Cho17~\cite{cho2017} and Lee19~\cite{lee2019}.

\begin{figure}[H]
    \centering
    \includegraphics[width=0.6\linewidth]{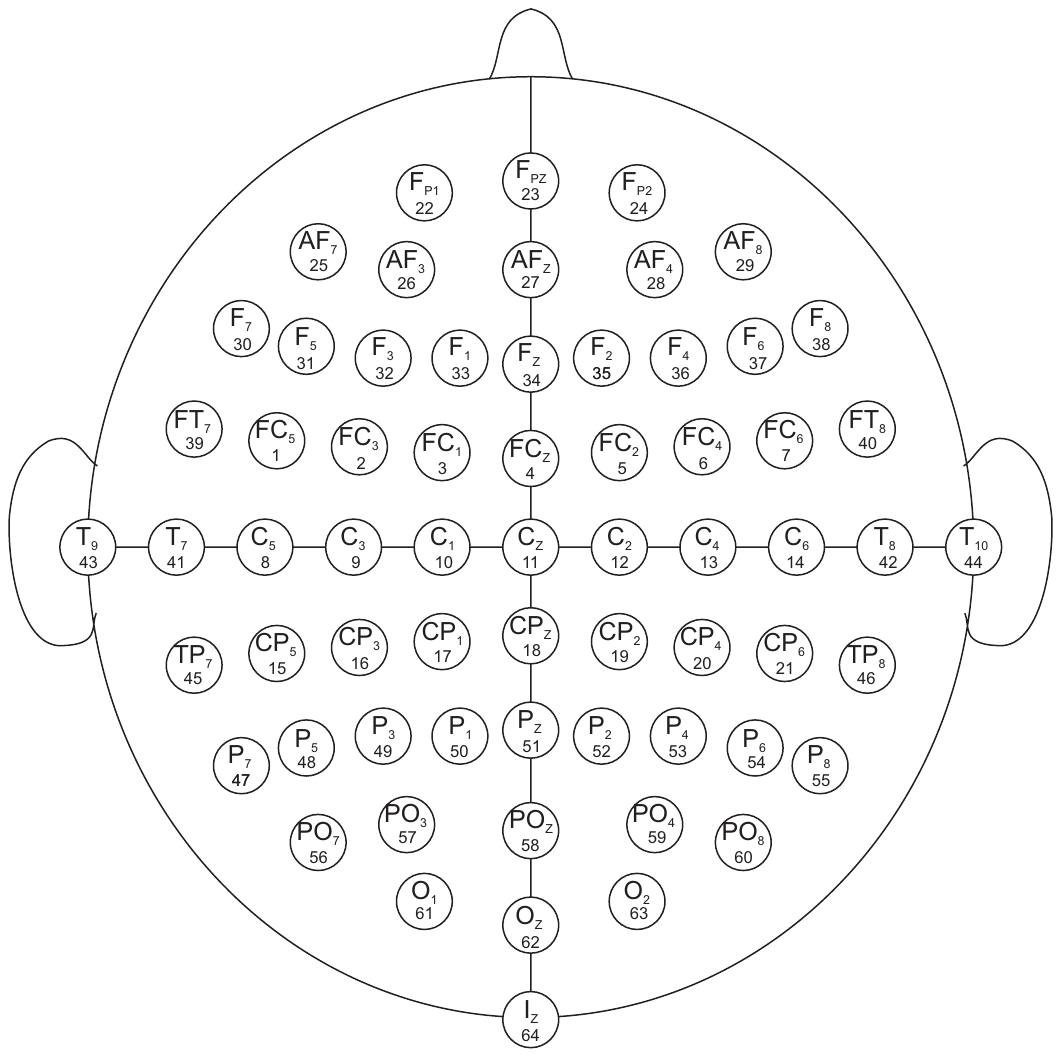}
    \caption{International 10-10 system EEG channel labeling.}
    \label{fig:sharbrough}
\end{figure}

The 48 channels are: AF7, CP5, AF4, P4, P8, P2, FC6, Fz, C5, O1, Fp1, Fp2, F4, CP4, PO3, C1, FC1, T8, Pz, Oz, TP7, Cz, FC2, CP6, CP2, POz, PO4, C6, P7, AF3, FC4, TP8, CP1, O2, C2, F8, FC3, P3, AF8, FC5, F7, F3, T7, C4, CP3, CPz, C3, P1. The channel locations are shown in Figure~\ref{fig:sharbrough}.

\paragraph{Preprocessing}
This section details the preprocessing steps taken for the PCL dataset. The raw PCL dataset contains data from 2 machines, Schalk04~\cite{physionetMI} and Cho17~\cite{cho2017} both used a BCI2000 system~\cite{physionetMI} while Lee19~\cite{lee2019} used an undefined machine. Both machines have different channels and sampling frequency characteristics. We take only the 48 channels they have in common and we resample the data to a standard sampling frequency of 250Hz for both machines. We then apply a bandpass filter from 0.3Hz to 30Hz. This bandpass filter removes frequency bands generally considered uninformative for the motor imagery task. Finally, we then detrend the data and normalize the three second recordings with a standard scaler applied to the channels individually.

\paragraph{Domain information}
Table~\ref{table:pcl_domain_participants} details the number of participants per domain and some demographic information; we put N/A for unavailable demographic information. Table~\ref{table:pcl_domain_proportions} details the proportion of samples and labels across domains.

\begin{table}[H]
    \centering
    \caption{Number of participants and demographic information of the PCL dataset}
    \begin{tabular}{lc|cc|c}
        \toprule
        \textbf{Domain} & Number of participants & Male & Female & Age  \\
        \midrule
        Schalk04     & 109 & N/A & N/A & N/A \\
        Cho2017         & 52 & 33 & 19 & $24.8\pm 3.9$ \\
        Lee19     & 54 & 29 & 25 & $[24,\,35]$ \\
        \midrule
        \textbf{Total} & 215 & N/A & N/A & N/A \\
        \bottomrule
    \end{tabular}
    \label{table:pcl_domain_participants}
\end{table}

\begin{table}[H]
    \centering
    \caption{Domain proportions of labels in the PCL dataset}
    \begin{tabular}{lccccccc}
        \toprule
        \textbf{Domain} & Left Hand & Right Hand & \textbf{Domain Total} \\
        \midrule
        Schalk04     & 2480  & 2438  & 4918 \\
        Cho2017         & 4940  & 4940  & 9880 \\
        Lee19         & 3900  & 3900  & 7800 \\
        \midrule
        \textbf{Total}  & 11320  & 11278 & 22598\\
        \bottomrule
    \end{tabular}
    \label{table:pcl_domain_proportions}
\end{table}

\paragraph{Architecture choice}
For this dataset, we use a deep convolution network model as defined in work from~\citet{eegnet}. We use the implementation of the BrainDecode~\citet{braindecode} Toolbox. We chose this model because it is well recognized by the EEG community. It also has a smaller architecture that better fits the data amount and task complexity of the PCL dataset. The implementation is available at \url{https://github.com/TNTLFreiburg/braindecode}.

\subsubsection{Detailed results}
\paragraph{ID evaluation}
We show the results of ERM for the PCL dataset in Table~\ref{table:pcl_ID}. We obtain these results by doing a hyperparameter search with the methodology detailed in Appendix~\ref{appendix:framework} with no held-out test domain and choose the model with train-domain validation. In other words, the training is done with all domains; thus, all domains are ID. The columns correspond to the validation accuracy of the chosen model in each domain. 

\begin{table}[H]
    \centering
    \caption{ID results for the PCL dataset}
    \adjustbox{max width=\textwidth}{%
        \begin{tabular}{lccccc}
            \toprule
            \textbf{Algorithm} & Schalk04 & Cho17 & Lee19 & \textbf{Average}\\
            \midrule
            ID ERM &  $76.40 \, (0.19)$  &  $68.07 \, (0.09)$  &  $76.45 \, (0.23)$  &  73.64 \\
            \bottomrule
        \end{tabular}
    }
    \label{table:pcl_ID}
\end{table}

\paragraph{Benchmark results}
We show the detailed benchmark results of the adapted OOD generalization algorithms in Table~\ref{table:detailed_pcl_results}. Each results is obtained by holding out one domain during training and reporting the performance of the chosen model from the hyperparameter sweep on that held out domain, more details in Appendix~\ref{appendix:framework}.

\begin{table*}[h]
	\centering
	\caption{OOD generalization algorithms performance on the PCL dataset}
	\begin{minipage}{\linewidth}
		\begin{center}
		\adjustbox{max width=\textwidth}{%
			\begin{tabular}{lcccc}
				\toprule
				\multicolumn{5}{c}{\textbf{Train-domain validation}} \\
				\midrule
				\textbf{Objective} & Schalk04 & Cho17 & Lee19 & \textbf{Average}\\
				\midrule
				ERM &  $63.52 \, (0.92)$  &  $59.34 \, (0.23)$  &  $70.06 \, (0.46)$  &  64.31 \\
				IRM &  $63.43 \, (0.37)$  &  $60.41 \, (0.07)$  &  $67.90 \, (0.27)$  &  63.91 \\
				VREx &  $62.65 \, (0.29)$  &  $58.84 \, (0.26)$  &  $68.22 \, (0.33)$  &  63.24 \\
				GroupDRO &  $63.97 \, (0.57)$  &  $60.24 \, (0.35)$  &  $70.34 \, (0.02)$  &  64.85 \\
                IB-ERM &  $63.31 \, (0.16)$  &  $59.82 \, (0.38)$  &  $70.18 \, (0.41)$  &  64.44 \\
                SD &  $63.72 \, (0.20)$  &  $59.31 \, (0.36)$  &  $70.15 \, (0.16)$  &  64.40 \\
				\bottomrule
			\end{tabular}}
		\end{center}
	\end{minipage}
	\begin{minipage}{\linewidth}
		\begin{center}
		\adjustbox{max width=\textwidth}{%
			\begin{tabular}{lcccc}
				\toprule
				\multicolumn{5}{c}{\textbf{Oracle train-domain validation}} \\
				\midrule
				\textbf{Objective} & Schalk04 & Cho17 & Lee19 & \textbf{Average}\\
				\midrule
				ERM &  $64.52 \, (0.25)$  &  $60.41 \, (0.22)$  &  $71.11 \, (0.29)$  &  65.35 \\
				IRM &  $63.28 \, (0.30)$  &  $61.09 \, (0.42)$  &  $68.77 \, (0.41)$  &  64.38 \\
				VREx &  $62.41 \, (0.47)$  &  $59.28 \, (0.29)$  &  $68.08 \, (0.20)$  &  63.26 \\
				GroupDRO &  $63.96 \, (0.36)$  &  $59.60 \, (0.38)$  &  $70.09 \, (0.16)$  &  64.55 \\
                IB-ERM &  $64.63 \, (0.22)$  &  $60.22 \, (0.38)$  &  $70.27 \, (0.34)$  &  65.04 \\
                SD &  $64.50 \, (0.05)$  &  $60.80 \, (0.17)$  &  $70.72 \, (0.14)$  &  65.34 \\
				\bottomrule
			\end{tabular}}
		\end{center}
	\end{minipage}
	\label{table:detailed_pcl_results}
\end{table*}

\subsubsection{Credit and license}
This dataset is built from 3 different motor imagery datasets~\cite{physionetMI,cho2017,lee2019} as made available on the online MOABB~\cite{moabb} platform. The PhysionetMI dataset is licensed under the Open Data Commons Attribution license v1.0.

\subsection{LSA64} \label{appendix:lsa64}

\begin{figure}[H]
    \centering
    \includegraphics[width=0.8\linewidth]{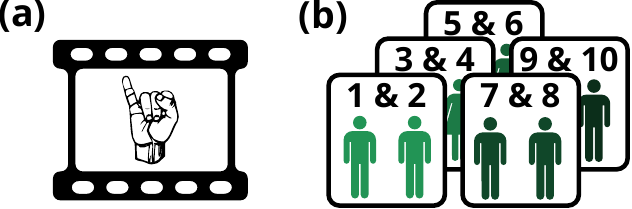}
    \caption{Summary of the LSA64 dataset. (a) The task is to perform signed word classification from videos. (b) The dataset has five source domains, where each domain contains videos of different signers. The goal is to generalize to unseen signers.}
    \label{fig:lsa64_appendix}
\end{figure}

\subsubsection{Setup}

\paragraph{Motivation}
Communication is an individualistic way to convey information through different media: text, speech, body language, and many others. However, some media are more distinctive and challenging than others. For example, text communication has less inter-individual variability than body language or speech. If deep learning systems hope to interact with humans effectively, models need to generalize to new and evolving mannerisms, accents, and other subtle variations in communication that significantly impact the meaning of the message conveyed. We study the ability of models to recognize information coming from unseen individuals with the LSA64~\cite{lsa64} dataset (Figure~\ref{fig:lsa64_appendix}).

\paragraph{Problem setting}
We consider the video classification of signed words in Argentinian Sign Language. The dataset has five source domains, where each domain contains videos of different signers. The goal is to generalize to unseen signers.

\paragraph{Data}
The dataset consists of 3200 videos from 10 different signers signing in Argentinian Sign Language. The inputs $\mathbf X$ are videos of 20 frames with resolution $(3,\,224,\,224)$. Sequences are two and a half seconds long. The labels $\mathbf Y$ consist of 64 words: Opaque, Red, Green, Yellow, Bright, Light-blue, Colors, Light-red Women, Enemy, Son, Man, Away, Drawer, Born, learn, Call, Skimmer, Bitter, Sweet milk, Milk, Water, Food, Argentina, Uruguay, Country, Last name, Where, Mock, Birthday, Breakfast, Photo, Hungry, Map, Coin, Music, Ship, None, Name, Patience, Perfume, Deaf, Trap, Rice, Barbecue, Cady, Chewing-gum, Spaghetti, Yogurt, accept, Thanks, Shut down, Appear, To land, Catch, Help, Dance, Bathe, Buy, Copy, Run, Realize, Give, and Find. The domains $d$ are 5 subgroups of signers: Signers 1 \& 2, Signers 3 \& 4, Signers 5 \& 6, Signers 7 \& 8 and Signers 9 \& 10.

\paragraph{Preprocessing}
This section details the preprocessing steps taken for the LSA64 dataset. The raw LSA64 dataset contains 3200 videos, each about 3 seconds long with the resolution of 1920x1080, at 60 frames per second. We first crop all videos at precisely 2.5 seconds to have videos of the same length. This cropping does not impact the information content of the video as signers pause at the end of their signed words. We then resize the frames to 224x224. Finally, we use PyTorchVideo~\cite{pytorchvideo} to uniformly sample 20 frames from each video in a sequence for prediction.

\paragraph{Domain information}
Table~\ref{table:lsa64_domain_proportions} details the proportion of samples and labels across domains.
\begin{table}[H]
    \centering
    \caption{Domain proportions of labels in the LSA64 dataset}
    \begin{tabular}{lccccccc}
        \toprule
        \textbf{Domain} & 64 words & \textbf{Domain Total} \\
        \midrule
        Signer 1 \& 2 & 10 videos per word  & 640 \\
        Signer 3 \& 4 & 10 videos per word  & 640 \\
        Signer 5 \& 6 & 10 videos per word  & 640 \\
        Signer 7 \& 8 & 10 videos per word  & 640 \\
        Signer 9 \& 10 & 10 videos per word  & 640 \\
        \midrule
        \textbf{Total}  & 50 videos per word & 3200\\
        \bottomrule
    \end{tabular}
    \label{table:lsa64_domain_proportions}
\end{table}

\paragraph{Architecture choice}
We use a Convolutional Recurrent Neural Network (CRNN) for this dataset. The CRNN model has 4 model blocks: Convolutional, Recurrent, attention, and prediction. First, we feed each video frame through a frozen Resnet50 model that is pretrained on Imagenet to extract relevant features. We then feed these feature vectors sequentially to an LSTM model. Finally, push the output of the LSTM model for each frame through a self-attention layer which linearly combines the LSTM output weighed by their attention scores. We the use a fully connected network to make predictions. Table~\ref{table:LSA64_arch} details the layers of the model architecture.

\begin{table}[H] 
    \centering
    \caption{Model architecture used for the LSA64 dataset}
    \begin{tabular}{cl}
        \toprule
        \textbf{\#} & Layer \\
        \midrule
        1 & Resnet50(in=3x224x224, out=2048)\\
        2 & Linear(in=2048, out=512)\\
        3 & ReLU\\
        4 & BatchNorm(num\_features=512, momentum=0.01)\\
        5 & Linear(in=512, out=512)\\
        6 & ReLU\\
        7 & BatchNorm(num\_features=512, momentum=0.01)\\
        8 & Linear(in=512, out=216)\\
        9 & ReLU\\
        10 & LSTM(in=256, hidden\_size=128, num\_layers=2)\\
        11 & SelfAttention(in=128, out=128)\\
        12 & Linear(in=128, out=64)\\
        13 & ReLU\\
        14 & Linear(in=64, out=64)\\
        \bottomrule
    \end{tabular}
    \label{table:LSA64_arch}
\end{table}

\subsubsection{Detailed Results}

\paragraph{ID evaluation}
We show the results of ERM for the LSA64 dataset in Table~\ref{table:lsa64_ID}. We obtain these results by doing a hyperparameter search with the methodology detailed in Appendix~\ref{appendix:framework} with no held-out test domain and choose the model with train-domain validation. In other words, the training is done with all domains; thus, all domains are ID. The columns correspond to the validation accuracy of the chosen model in each domain. 
\begin{table}[H]
    \centering
    \caption{ID results for the LSA64 dataset}
    \adjustbox{max width=\textwidth}{%
        \begin{tabular}{lccccccc}
            \toprule
            \textbf{Algorithm} & Signers 1 \& 2 & Signers 3 \& 4 & Signers 5 \& 6 & Signers 7 \& 8 & Signers 9 \& 10 & \textbf{Average}\\
            \midrule
            ID ERM &  $90.10 \, (0.56)$  &  $89.58 \, (1.13)$  &  $80.21 \, (1.06)$  &  $85.16 \, (1.61)$  &  $87.76 \, (0.77)$  &  86.56 \\
            \bottomrule
        \end{tabular}
    }
    \label{table:lsa64_ID}
\end{table}

\paragraph{Benchmarks results}
We show the detailed benchmark results of the adapted OOD generalization algorithms in Table~\ref{table:detailed_lsa64_results}. Each results is obtained by holding out one domain during training and reporting the performance of the chosen model from the hyperparameter sweep on that held out domain, more details in Appendix~\ref{appendix:framework}.

\begin{table*}[h]
	\centering
	\caption{OOD generalization algorithms performance on the LSA64 dataset}
	\begin{minipage}{\linewidth}
		\begin{center}
		\adjustbox{max width=\textwidth}{%
			\begin{tabular}{lcccccc}
				\toprule
				\multicolumn{7}{c}{\textbf{Train-domain validation}} \\
				\midrule
				\textbf{Objective} & Signers 1 \& 2 & Signers 3 \& 4 & Signers 5 \& 6 & Signers 7 \& 8 & Signers 9 \& 10 & \textbf{Average}\\
				\midrule
				ERM &  $48.50 \, (2.93)$  &  $50.65 \, (2.28)$  &  $47.53 \, (1.32)$  &  $57.49 \, (2.49)$  &  $62.96 \, (0.98)$  &  53.42 \\
				IRM &  $44.34 \, (0.60)$  &  $43.16 \, (1.48)$  &  $38.28 \, (2.01)$  &  $46.88 \, (1.13)$  &  $52.47 \, (2.78)$  &  45.03 \\
				VREx &  $42.19 \, (3.64)$  &  $45.57 \, (1.67)$  &  $42.06 \, (2.82)$  &  $51.82 \, (1.95)$  &  $52.21 \, (4.26)$  &  46.77 \\
				GroupDRO &  $43.62 \, (3.95)$  &  $44.14 \, (1.87)$  &  $43.29 \, (1.58)$  &  $47.79 \, (1.70)$  &  $52.73 \, (1.36)$  &  46.32 \\
                IB-ERM &  $55.66 \, (1.71)$  &  $56.71 \, (2.16)$  &  $49.80 \, (2.41)$  &  $64.52 \, (0.61)$  &  $59.70 \, (2.51)$  &  57.28 \\
                SD &  $48.63 \, (2.46)$  &  $50.20 \, (1.60)$  &  $40.89 \, (0.84)$  &  $57.68 \, (2.54)$  &  $56.32 \, (1.19)$  &  50.74 \\
				\bottomrule
			\end{tabular}}
		\end{center}
	\end{minipage}
	\begin{minipage}{\linewidth}
		\begin{center}
		\adjustbox{max width=\textwidth}{%
			\begin{tabular}{lcccccc}
				\toprule
				\multicolumn{7}{c}{\textbf{Oracle train-domain validation}} \\
				\midrule
				\textbf{Objective} & Signers 1 \& 2 & Signers 3 \& 4 & Signers 5 \& 6 & Signers 7 \& 8 & Signers 9 \& 10 & \textbf{Average}\\
				\midrule
				ERM &  $54.43 \, (1.11)$  &  $59.24 \, (0.23)$  &  $48.89 \, (1.45)$  &  $62.96 \, (1.25)$  &  $65.62 \, (0.42)$  &  58.23 \\
				IRM &  $42.06 \, (1.17)$  &  $43.16 \, (1.48)$  &  $39.06 \, (1.39)$  &  $46.22 \, (3.83)$  &  $47.46 \, (1.95)$  &  43.59 \\
				VREx &  $46.29 \, (1.81)$  &  $49.93 \, (0.45)$  &  $42.84 \, (0.75)$  &  $54.23 \, (0.59)$  &  $56.90 \, (0.53)$  &  50.04 \\
				GroupDRO &  $50.52 \, (1.01)$  &  $54.49 \, (1.87)$  &  $45.12 \, (1.27)$  &  $56.51 \, (1.66)$  &  $63.22 \, (0.75)$  &  53.97 \\
                IB-ERM &  $56.51 \, (1.38)$  &  $59.51 \, (0.91)$  &  $51.82 \, (0.96)$  &  $64.52 \, (0.61)$  &  $66.54 \, (1.27)$  &  59.78 \\
                SD &  $56.58 \, (1.24)$  &  $60.68 \, (1.08)$  &  $49.35 \, (0.51)$  &  $62.43 \, (0.83)$  &  $64.06 \, (1.39)$  &  58.62 \\
				\bottomrule
			\end{tabular}}
		\end{center}
	\end{minipage}
	\label{table:detailed_lsa64_results}
\end{table*}

\subsubsection{Credit and license}
This dataset was adapted from the work of~\citet{lsa64}. The LSA64 dataset is under the Creative Commons Attribution-NonCommercial-ShareAlike 4.0 International License.

\subsection{HHAR} \label{appendix:hhar}

\begin{figure}[H]
    \centering
    \includegraphics[width=0.8\linewidth]{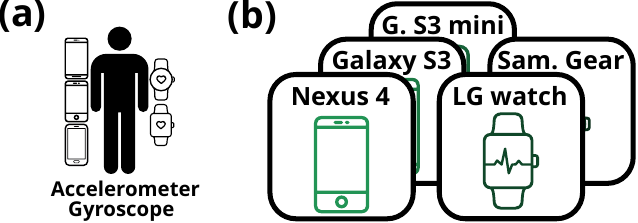}
    \caption{Summary of the HHAR dataset. (a) The task is to perform human activity classification from smart devices sensory data. (b) The dataset has five source domains, where each domain contains data gathered with a different smart device. The goal is to generalize to unseen smart devices.}
    \label{fig:hhar_appendix}
\end{figure}

\subsubsection{Setup}

\paragraph{Motivation}
The intrinsic biases from inaccurate and poorly calibrated sensors of smart devices, along with the accumulated biases from everyday use makes human activity recognition a notoriously difficult task when task when done across devices~\cite{hhar,blunck2013heterogeneity}. Contrary to static tasks where uninformative features can often be segmented out from the input features (e.g., background when classifying an animal from an image), invariant features in time series are often highly convoluted with other spurious features. We study the ability of models to ignore spurious information from complex signals with the HHAR~\cite{hhar,uci} dataset (Figure~\ref{fig:hhar_appendix}).

\paragraph{Problem setting}
We consider the human activity classification task from accelerometer and gyroscope measurements of smartphones and smartwatches. The dataset has five source domains, where each domain contains data gathered with a different device. The goal is to generalize to unseen smart devices.

\paragraph{Data}
The dataset consists of 13674 recordings of 3-axis accelerometer and 3-axis gyroscope data from 5 different smart devices (3 smartphones and 2 smartwatches). The inputs $\mathbf X$ are five second recordings of a 6-dimensional signal sampled at 100Hz. The labels $\mathbf Y$ consist of 6 activities: Stand, Sit, Walk, Bike, Stairs up, and Stairs Down. Domains $d$ consist of five smart device models: Nexus 4, Galaxy S3, Galaxy S3 Mini, LG Watch, and Samsung Galaxy Gears.

\paragraph{Preprocessing}
This section details the preprocessing steps taken for the HHAR dataset. The raw data was gathered with 10 different smart devices (2 from each model). Different models have different sampling frequencies, plus recordings have gaps in the data samples where devices temporarily stopped recording, making the time series irregularly sampled. We first remove the recordings of any device that either is missing considerable amounts of signals or has less than 100 seconds of recording. We then sort the data points in each sequence according to their recorded time, instead of time the data was saved on the device. Next, we split the full recordings into sequences of five seconds and resample at 100Hz. Finally, we normalize the data with a standard scaler applied to the accelerometer and gyroscope channels separately.

\paragraph{Domain information}
Table~\ref{table:hhar_domain_proportions} details the proportion of samples and labels across domains.

\begin{table}[H]
    \centering
    \caption{Domain proportions of labels in the HHAR dataset}
    \begin{tabular}{lccccccc}
        \toprule
        \textbf{Domain} & Stand & Sit & Walk & Bike & Stairs up & Stairs down & \textbf{Domain Total} \\
        \midrule
        Nexus 4         & 760   & 911   &  1024     &  644   &  695   & 543  & 4577 \\
        Galaxy S3       & 664   & 889   &  944      &  560   &  635   & 474  & 4166 \\
        Galaxy S3 Mini  & 409   & 501   &  524      &  297   &  396   & 280  & 2407 \\
        LG watch        & 368   & 358   &  382      &  424   &  315   & 307  & 2154 \\
        Gear watch      & 21    & 23    &  78       &  42    &  120   & 86   & 370 \\
        \midrule
        \textbf{Total}  & 2222  & 2682  &  2952     &  1967  &  2161  & 1690 & 13674\\
        \bottomrule
    \end{tabular}
    \label{table:hhar_domain_proportions}
\end{table}

\paragraph{Architecture choice}
As this data is similar to EEG recordings, we use the same deep convolution network model as in the CAP and SEDFx datasets. The architecture is defined in work from~\citet{braindecode}. We use the implementation of the BrainDecode~\cite{braindecode} Toolbox. Temporal Convolutional Networks (TCN) are powerful tools for processing time series data~\cite{tcn}. The architecture we use combines temporal and spatial convolution, which fits this data well. We found that it performed well on this task and obtained stable performance. The implementation is available at \url{https://github.com/TNTLFreiburg/braindecode}.

\subsubsection{Detailed results}

\paragraph{ID evaluation}
We show the results of ERM for the HHAR dataset in Table~\ref{table:hhar_ID}. We obtain these results by doing a hyperparameter search with the methodology detailed in Appendix~\ref{appendix:framework} with no held-out test domain and choose the model with train-domain validation. In other words, the training is done with all domains; thus, all domains are ID. The columns correspond to the validation accuracy of the chosen model in each domain. 
\begin{table}[H]
    \centering
    \caption{ID results for the HHAR dataset}
    \adjustbox{max width=\textwidth}{%
        \begin{tabular}{lccccccc}
            \toprule
            \textbf{Algorithm} & Nexus 4 & Galazy S3 & Galaxy S3 Mini & LG watch & Sam. Gear & \textbf{Average}\\
            \midrule
            ID ERM &  $98.91 \, (0.24)$  &  $98.44 \, (0.15)$  &  $98.68 \, (0.15)$  &  $90.08 \, (0.28)$  &  $80.63 \, (1.33)$  &  93.35 \\
            \bottomrule
        \end{tabular}
    }
    \label{table:hhar_ID}
\end{table}

\paragraph{Benchmark results}
We show the detailed benchmark results of the adapted OOD generalization algorithms in Table~\ref{table:detailed_hhar_results}. Each results is obtained by holding out one domain during training and reporting the performance of the chosen model from the hyperparameter sweep on that held out domain, more details in Appendix~\ref{appendix:framework}.

\begin{table*}[h]
	\centering
	\caption{OOD generalization algorithms performance on the HHAR dataset}
	\begin{minipage}{\linewidth}
		\begin{center}
		\adjustbox{max width=\textwidth}{%
			\begin{tabular}{lcccccc}
				\toprule
				\multicolumn{7}{c}{\textbf{Train-domain validation}} \\
				\midrule
				\textbf{Objective} & Nexus 4 & Galazy S3 & Galaxy S3 Mini & LG watch & Sam. Gear & \textbf{Average}\\
				\midrule
				ERM &  $97.91 \, (0.03)$  &  $98.17 \, (0.18)$  &  $92.49 \, (0.26)$  &  $71.33 \, (0.67)$  &  $62.16 \, (1.69)$  &  84.41 \\
				IRM &  $95.68 \, (0.47)$  &  $96.31 \, (0.53)$  &  $91.10 \, (0.35)$  &  $69.76 \, (1.44)$  &  $61.71 \, (1.56)$  &  82.91 \\
				VREx &  $95.53 \, (0.55)$  &  $96.51 \, (0.16)$  &  $91.36 \, (0.43)$  &  $69.72 \, (0.29)$  &  $62.73 \, (1.15)$  &  83.17 \\
				GroupDRO &  $96.49 \, (0.18)$  &  $96.79 \, (0.12)$  &  $92.13 \, (0.09)$  &  $71.64 \, (0.43)$  &  $63.74 \, (1.34)$  &  84.16 \\
                IB-ERM &  $97.56 \, (0.06)$  &  $97.93 \, (0.21)$  &  $91.76 \, (0.57)$  &  $71.38 \, (1.02)$  &  $59.01 \, (1.86)$  &  83.53 \\
                SD &  $98.14 \, (0.01)$  &  $98.32 \, (0.19)$  &  $92.71 \, (0.09)$  &  $75.12 \, (0.18)$  &  $63.85 \, (0.28)$  &  85.63 \\
				\bottomrule
			\end{tabular}}
		\end{center}
	\end{minipage}
	\begin{minipage}{\linewidth}
		\begin{center}
		\adjustbox{max width=\textwidth}{%
			\begin{tabular}{lcccccc}
				\toprule
				\multicolumn{7}{c}{\textbf{Oracle train-domain validation}} \\
				\midrule
				\textbf{Objective} & Nexus 4 & Galazy S3 & Galaxy S3 Mini & LG watch & Sam. Gear & \textbf{Average}\\
				\midrule
				ERM &  $97.64 \, (0.06)$  &  $98.05 \, (0.07)$  &  $93.18 \, (0.20)$  &  $73.11 \, (0.77)$  &  $64.64 \, (1.20)$  &  85.32 \\
				IRM &  $96.81 \, (0.14)$  &  $96.43 \, (0.09)$  &  $91.26 \, (0.23)$  &  $70.61 \, (0.51)$  &  $61.82 \, (2.21)$  &  83.39 \\
				VREx &  $96.60 \, (0.24)$  &  $96.68 \, (0.29)$  &  $92.00 \, (0.65)$  &  $71.67 \, (0.84)$  &  $59.23 \, (1.17)$  &  83.24 \\
				GroupDRO &  $96.54 \, (0.23)$  &  $96.94 \, (0.15)$  &  $91.62 \, (0.34)$  &  $71.33 \, (0.68)$  &  $64.86 \, (0.69)$  &  84.26 \\
                IB-ERM &  $98.16 \, (0.09)$  &  $98.22 \, (0.09)$  &  $93.18 \, (0.16)$  &  $73.40 \, (0.68)$  &  $64.64 \, (0.09)$  &  85.52 \\
                SD &  $98.48 \, (0.01)$  &  $98.67 \, (0.11)$  &  $94.36 \, (0.24)$  &  $75.12 \, (0.18)$  &  $64.86 \, (0.28)$  &  86.30 \\
				\bottomrule
			\end{tabular}}
		\end{center}
	\end{minipage}
	\label{table:detailed_hhar_results}
\end{table*}

\subsubsection{Credits and license}

This dataset was adapted from the work of~\citet{hhar} as made available on the online UCI Machine Learning Repository~\cite{uci}. This dataset is licensed under the Open Data Commons Attribution license v1.0.

\subsection{AusElec} \label{appendix:auselec}

\begin{figure}[H]
    \centering
    \includegraphics[width=0.8\linewidth]{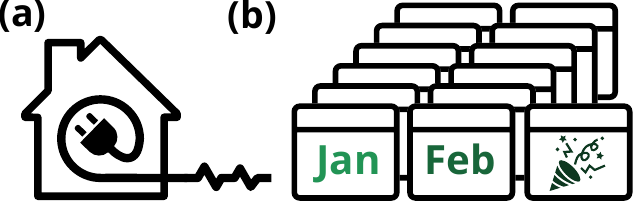}
    \caption{Summary of the AusElec dataset. (a) The task is to forecast electricity consumption. (b) The dataset has 13 time domains, where each domain contains data from different months and holidays. The goal is to perform well on all seasonalities.}
    \label{fig:auselec_appendix}
\end{figure}

\subsubsection{Setup}
\paragraph{Motivation}
Seasonality is the property of time series where recurring characteristics appear every cycle of a fixed period, e.g., weekly. A common practice in the forecasting field is to provide models with additional information, e.g., day of week in order to allow models to leverage seasonality for better predictions. However, holidays is a seasonality of time series that is very sparse which models often fail to capture. We study the performance of models on sparse seasonality with the AusElec~\cite{auselec,monash} dataset (Figure~\ref{fig:auselec_appendix})

\paragraph{Problem setting}
We consider the electricity consumption forecasting task. The dataset has 13 time domains, where each domain contains data from different months and holidays. The goal is to perform well on all seasonalities. 

\paragraph{Data}
The dataset consists of five time series comprising 13 years of electricity demand across five states in Australia: Victoria, New South Wales, Queensland, Tasmania and South Australia. The inputs $\mathbf X$ are seven days of electricity demand sampled half hourly to which we add 42 lag features and 5 time features. Lag features are past electricity demand values the goes past the seven day context given to the model. The time features are time indicators: minute of hour, hour of day, day of week, day of month, and day of year. The labels $\mathbf Y$ is the electricity demand for the day following the seven days of context. Domains $d$ consist of time intervals throughout the year: January, February, March, April, May, June, July, August, September, October, November, December, and holidays.

\paragraph{Preprocessing}
We do not perform any preprocessing for this dataset, this was already accomplished by prior work from~\citet{monash}.

\paragraph{Domain information}
We define the time interval of the Holidays domain as union of the following Australian holidays: New Year's Day, Australia Day, Good Friday, Easter Monday, Anzac Day, Christmas Day, Boxing Day.

\paragraph{Architecture choice}
For this dataset, we use a forecasting Transformer architecture closely following the original formulation of~\citet{attnisallyouneed}. We found that it performed well on this task and obtained stable performance. Details are in Table~\ref{table:auselec_arch}.

\begin{table}[H] 
    \centering
    \caption{Model architecture used for the AusElec dataset}
    \begin{tabular}{cl}
        \toprule
        \textbf{\#} & Layer \\
        \midrule
        1 & TransformerEncoder(d\_model=48, nhead=2, num\_encoder\_layers=2,\\
          &  \hspace{3cm} dim\_feedforward=32, dropout=0.1, activation=gelu)\\
        2 & TransformerDecoder(d\_model=48, nhead=2, num\_encoder\_layers=2, \\
          & \hspace{3cm} dim\_feedforward=32, dropout=0.1, activation=gelu)\\
        \bottomrule
    \end{tabular}
    \label{table:auselec_arch}
\end{table}

\subsubsection{Detailed Results}

\paragraph{Unbalanced results}
It has been reported in prior work~\cite{wilds} that OOD generalization algorithms such as IRM outperforms ERM on subpopulation shift datasets. However, it is unclear whether the improvements originates from the nature OOD generalization algorithms to upsample minority domains when computing the empirical risk or because the algorithm is performing well. In this work, we create an Unbalanced dataset of the subpopulation shift dataset which is agnostic of the domain definition during training. This allows us to compare the gain in performance obtained by upsampling the minority domain when minimizing the empirical risk. We show those results in Table~\ref{table:unbalanced_auselec_results}.
\begin{table*}[h]
	\centering
	\caption{Results for the AusElecUnbalanced dataset}
	\begin{minipage}{0.49\linewidth}
		\begin{center}
		\adjustbox{max width=\linewidth}{%
			\begin{tabular}{lcc}
				\toprule
				\multicolumn{3}{c}{\textbf{Average validation}} \\
				\midrule
				\textbf{Objective} & Average & \textbf{Worse}\\
				\midrule
				ERM &  $227.73 \, (2.64)$  &  $409.80 \, (4.21)$ \\
				\bottomrule
			\end{tabular}}
		\end{center}
	\end{minipage}
	\begin{minipage}{0.49\linewidth}
		\begin{center}
		\adjustbox{max width=\linewidth}{%
			\begin{tabular}{lcc}
				\toprule
				\multicolumn{3}{c}{\textbf{Worst-domain validation}} \\
				\midrule
				\textbf{Objective} & Average & \textbf{Worse}\\
				\midrule
				ERM &  $235.40 \, (4.38)$  &  $395.99 \, (5.49)$ \\
				\bottomrule
			\end{tabular}}
		\end{center}
	\end{minipage}
	\label{table:unbalanced_auselec_results}
\end{table*}

\paragraph{Benchmark results}
We show the detailed benchmark results of the adapted OOD generalization algorithms in Table~\ref{table:detailed_auselec_results}. Each line is obtained by training on all domains of the dataset and reporting the average and worst domain performance of the chosen model, more details in Appendix~\ref{appendix:framework}.
\begin{table*}[h]
	\centering
	\caption{OOD generalization algorithms performance on the AusElectricity dataset}
	\begin{minipage}{0.49\linewidth}
		\begin{center}
		\adjustbox{max width=\linewidth}{%
			\begin{tabular}{lcc}
				\toprule
				\multicolumn{3}{c}{\textbf{Average validation}} \\
				\midrule
				\textbf{Objective} & Average & \textbf{Worse}\\
				\midrule
				ERM &  $232.01 \, (2.60)$  &  $397.27 \, (8.48)$ \\
				VREx &  $237.96 \, (2.53)$  &  $415.01 \, (9.92)$ \\
				GroupDRO &  $237.09 \, (3.63)$  &  $408.83 \, (2.37)$ \\
				IB-ERM &  $232.03 \, (2.68)$  &  $393.56 \, (2.41)$ \\
				\bottomrule
			\end{tabular}}
		\end{center}
	\end{minipage}
	\begin{minipage}{0.49\linewidth}
		\begin{center}
		\adjustbox{max width=\linewidth}{%
			\begin{tabular}{lcc}
				\toprule
				\multicolumn{3}{c}{\textbf{Worst-domain validation}} \\
				\midrule
				\textbf{Objective} & Average & \textbf{Worse}\\
				\midrule
				ERM &  $247.08 \, (7.59)$  &  $403.56 \, (6.57)$ \\
				VREx &  $247.09 \, (2.19)$  &  $408.87 \, (3.97)$ \\
				GroupDRO &  $252.95 \, (7.58)$  &  $424.44 \, (13.34)$ \\
				IB-ERM &  $235.87 \, (3.11)$  &  $391.13 \, (5.44)$ \\
				\bottomrule
			\end{tabular}}
		\end{center}
	\end{minipage}
	\label{table:detailed_auselec_results}
\end{table*}

\subsubsection{Credits and license}

This dataset was adapted from the work of~\citet{auselec} as made available on the online Monash time series archive~\cite{monash}. This dataset is licensed under the Creative Commons Attribution 4.0 International License.

\subsection{IEMOCAP} \label{appendix:iemocap}

\begin{figure}[H]
    \centering
    \includegraphics[width=0.8\linewidth]{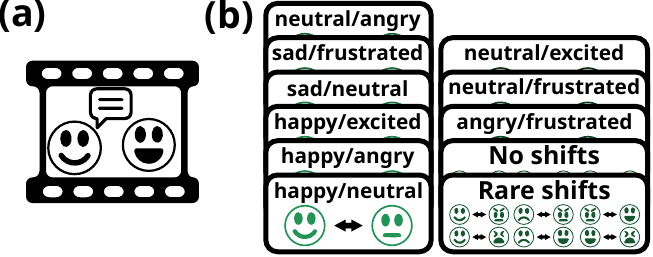}
    \caption{Summary of the IEMOCAP dataset. (a) The task is to perform emotion recognition from multi modal data (video, sound, text). (b) The dataset has 11 time domains, where each domain contains data from a different emotion shifts during conversations. The goal is to perform well on all conversational emotion shifts.}
    \label{fig:iemocap_description}
\end{figure}

\subsubsection{Setup}

\paragraph{Motivation}
Speakers tend to maintain an emotional state over a conversation. However, external stimuli can invoke a shift in the emotional state of speakers~\cite{poria2019emotion}. Such emotion shift are often sparsely represented in the data, making it hard for models to classify them adequately. Recent work on emotion recognition models~\cite{poria2019emotion,poria2018meld,majumder2019dialoguernn} show the failure of existing models to adapt to those emotion shift. We study the performance of models on emotional shift with the IEMOCAP~\cite{busso2008iemocap} dataset (Figure~\ref{fig:iemocap_main_body}).

\paragraph{Problem setting}
We consider the emotion recognition task. The dataset has 11 time domains, where each domain contains data from a different emotion shift during conversations. The goal is to perform well on all conversational emotion shifts.

\paragraph{Data}
The dataset consists of 151 videos about dyadic interactions, where professional actors are required to perform scripted scenes that elicit specific emotions. Each video contains a single dyadic dialogue, segmented into utterances. It contains 7433 utterances in total. The inputs $\mathbf X$ are utterances of  video, speech, and text transcriptions. The labels $\mathbf Y$ consist of 6 emotions: Happy, Sad, Neutral, Angry, Excited, and Frustrated. Domains $d$ consist of 11 emotion shift during conversations: No-Shift, Rare-Shift, and 9 common emotion shifts including Happy-Neutral, Happy-Angry, Happy-Excited, Sad-Neutral, Sad-Frustrated, Neutral-Angry, Neutral-Excited, Neutral-Frustrated, and Angry-Frustrated .

\paragraph{Preprocessing}
This section details the preprocessing steps taken to bring the IEMOCAP dataset from its raw form to its final form used in WOODS. For each utterance, we extract multimodal features (audio, visual and text) following the same approach as \citet{hazarika2018conversational} and \citet{majumder2019dialoguernn}. To get our text embedding, we use a simple CNN with one convolutional layer followed by max-pooling~\cite{poria2016deeper}. To extract high dimensional audio vectors, we use openSMILE~\cite{eyben2010opensmile}. These vectors comprise features like loudness, Mel-spectra, MFCC, pitch, etc. We use a 3D-CNN to capture video embeddings ~\cite{tran2015learning}. This embedding contains information for detecting emotional expressions like a smile or frown. We use concatenation of the unimodal features as
a fusion method.

\paragraph{Domain information}
We consider utterances that have the same label as the previous utterance spoken by the same speaker as a no-shift domain. We consider emotion-shifts that appear in less than 20 utterances as rare-shift domain, namely, Happy-Angry, Excited-Angry, Frustrated-Happy, Sad-Excited, frustrated-Excited, and Sad-Angry. We consider the remaining 9 emotion shifts as common ones and create a separate domain for each of them. For brevity, we call these domains common-shift in general. The ratios for the rare emotion-shift domain are 1/6, 1/6, and 2/3 for training, validation, and test respectively. For the remaining domains, dialogues are randomly chosen to achieve the ratios of 0.7, 0.1, and 0.2 for the size of training, validation, and test respectively.

Table~\ref{table:iemocap_domain_proportions} details the proportion of utterances and dialogues in the training, validation, and test sets across domains.

\begin{table}[H]
    \centering
    \caption{Domain proportions of utterances and dialogues in the training, validation and test sets of IEMOCAP dataset}
    \begin{tabular}{lccc}
        \toprule
            & Training & Validation & Test \\
        \midrule
        \# of utterances in rare-shift domain    & 22   & 19   &  61     \\
        \# of utterances in  no-shift domain      & 3785   & 369   &  957       \\
        total \# of utterance in common-shift domains  & 1297   & 196   &  527      \\
        \midrule
        \textbf{total \# of utterances}  & 5298  & 589  &  1546     \\
        \textbf{total \# of dialogues}  & 108  & 12  &  31 \\
        \bottomrule
    \end{tabular}
    \label{table:iemocap_domain_proportions}
\end{table}

\paragraph{Architecture choice}
For this dataset, we use a DialogueRNN  model as defined in work from~\citet{majumder2019dialoguernn}. We chose this model because it is well recognized by the ERM community. It also  has an  effective mechanisms to model context by tracking individual speaker states throughout the conversation for emotion classification. The implementation is available at \url{https://github.com/declare-lab/conv-emotion/tree/master/DialogueRNN}.

\subsubsection{Detailed results}

\paragraph{Unbalanced results}
It has been reported in prior work~\cite{wilds} that OOD generalization algorithms such as IRM outperforms ERM on subpopulation shift datasets. However, it is unclear whether the improvements originates from the nature OOD generalization algorithms to upsample minority domains when computing the empirical risk or because the algorithm is performing well. In this work, we create an Unbalanced dataset of the subpopulation shift dataset which is agnostic of the domain definition during training. This allows us to compare the gain in performance obtained by upsampling the minority domain when minimizing the empirical risk. We show those results in Table~\ref{table:unbalanced_iemocap_results}.
\begin{table*}[h]
	\centering
	\caption{Results for the IEMOCAPUnbalanced dataset}
	\begin{minipage}{0.49\linewidth}
		\begin{center}
		\adjustbox{max width=\linewidth}{%
			\begin{tabular}{lcc}
				\toprule
				\multicolumn{3}{c}{\textbf{Average validation}} \\
				\midrule
				\textbf{Objective} & Average & \textbf{Worst}\\
				\midrule
                ERM &  $70.53 \, (0.05)$  &  $58.24 \, (1.41)$ \\
				\bottomrule
			\end{tabular}}
		\end{center}
	\end{minipage}
	\begin{minipage}{0.49\linewidth}
		\begin{center}
		\adjustbox{max width=\linewidth}{%
			\begin{tabular}{lcc}
				\toprule
				\multicolumn{3}{c}{\textbf{Worst-domain validation}} \\
				\midrule
				\textbf{Objective} & Average & \textbf{Worst}\\
				\midrule
                ERM &  $70.01 \, (0.77)$  &  $56.76 \, (1.24)$ \\
				\bottomrule
			\end{tabular}}
		\end{center}
	\end{minipage}
	\label{table:unbalanced_iemocap_results}
\end{table*}

\paragraph{Benchmark results}
We show the detailed benchmark results of the adapted OOD generalization algorithms in Table~\ref{table:detailed_iemocap_results}. Each line is obtained by training on all domains of the dataset and reporting the average and worst domain performance of the chosen model, more details in Appendix~\ref{appendix:framework}.
\begin{table*}[h]
	\centering
	\caption{OOD generalization algorithms performance on the IEMOCAP dataset}
	\begin{minipage}{0.49\linewidth}
		\begin{center}
		\adjustbox{max width=\linewidth}{%
			\begin{tabular}{lcc}
				\toprule
				\multicolumn{3}{c}{\textbf{Average validation}} \\
				\midrule
				\textbf{Objective} & Average & \textbf{Worst}\\
				\midrule
                ERM &  $69.12 \, (0.36)$  &  $57.75 \, (1.85)$ \\
                IRM &  $68.73 \, (0.24)$  &  $55.93 \, (1.20)$ \\
                VREx &  $70.12 \, (0.51)$  &  $59.45 \, (1.43)$ \\
                GroupDRO &  $69.21 \, (0.75)$  &  $56.11 \, (1.19)$ \\
                IB-ERM &  $68.79 \, (0.08)$  &  $59.93 \, (0.55)$ \\
                SD &  $68.62 \, (0.22)$  &  $58.04 \, (0.39)$ \\
				\bottomrule
			\end{tabular}}
		\end{center}
	\end{minipage}
	\begin{minipage}{0.49\linewidth}
		\begin{center}
		\adjustbox{max width=\linewidth}{%
			\begin{tabular}{lcc}
				\toprule
				\multicolumn{3}{c}{\textbf{Worst-domain validation}} \\
				\midrule
				\textbf{Objective} & Average & \textbf{Worst}\\
				\midrule
                ERM &  $69.85 \, (0.03)$  &  $56.33 \, (2.76)$ \\
                IRM &  $70.21 \, (0.31)$  &  $58.95 \, (1.13)$ \\
                VREx &  $69.64 \, (0.44)$  &  $57.66 \, (3.13)$ \\
                GroupDRO &  $70.08 \, (0.86)$  &  $58.79 \, (1.00)$ \\
                IB-ERM &  $70.04 \, (0.42)$  &  $58.81 \, (1.50)$ \\
                SD &  $68.75 \, (0.28)$  &  $56.14 \, (1.24)$ \\
				\bottomrule
			\end{tabular}}
		\end{center}
	\end{minipage}
	\label{table:detailed_iemocap_results}
\end{table*}

\subsubsection{Credits and license}

This dataset was adapted from the work of~\citet{busso2008iemocap} as made available by the Speech Analysis and Interpretation
Laboratory (SAIL) at the University of Southern California (USC). This dataset is licensed under the license availabel at \url{https://sail.usc.edu/iemocap/iemocap_release.htm}.

\section{Further details on adapation of OOD generalization algorithms} \label{appendix:adaptation}

\subsection{General adaptation of OOD generalization algorithms to time series}
The problem formulation in Section~\ref{sect:problem_formulation_time_series} applies only sequence of same length $S_t$ and prediction times $S_p$ across samples. However, for several dataset and tasks, this does not hold up. Take as example the IEMOCAP dataset, conversations can vary in length and prediction times across samples. In this section, we provide a general formulation that accounts these changes.

Data samples consist of the input time series observation $\mathbf X^i = [X^i_{t}]_{t\in S^i_t}$, where $S^i_t$ is the set of time steps for sample $i$, and the set of labels $\mathbf Y^i = [Y^i_t]_{t\in S^i_p}$, where $S^i_p \subseteq S^i_t$ is the set of labeled time steps for sample $i$. 

\paragraph{Empirical risk}
For the empirical risk of domain $d$, we average the risk across the set of labeled time steps of sample $i$ belonging to domain $d$: $S_p^{d,i}$.
\begin{equation} \label{eq:general_emp_risk}
    R^d(f) = \frac{1}{n^d}\sum_{(\mathbf X^i, \mathbf Y^i)\in D} \frac{1}{|S_p^{d,i}|}\sum_{t\in S_p^{d,i}} \mathcal L\big(f(X^i_{1:t}),Y^i_t\big),
\end{equation}
where $n^d$ is the number of samples from domain $d$ in the dataset $D$. 

\paragraph{Penalty value function} 
IB-ERM and SD penalize representation and logits during prediction, we follow Equation (\ref{eq:emp_risk}) and define the penalty below.
\begin{equation} \label{eq:general_pen_val}
    P(f) = \frac{1}{n^d} \sum_{(\mathbf X^i, \mathbf Y^i)\in D} \frac{1}{|S_p^{d,i}|}\sum_{t\in S_p^{d,i}} \tilde P(f,X^i_{1:t}, Y^i_t).
\end{equation}

\subsection{OOD generalization algorithm definition}
\begin{itemize}
    \item \textbf{IRM} performs a constrained empirical risk minimization such that the optimal classifier of representations is the same across the domains. It does so by penalizing a function of empirical risk across domains. We adapt IRM by using the empirical risk from Equation (\ref{eq:general_emp_risk}):
        \begin{equation} \label{eq:irm_pen_val} \tag{IRM}
            P(f) = \frac{1}{d} \sum_{D^d\in D} \| \nabla_{w|w=1.0} R^d(w\cdot f) \|^2,
        \end{equation}
    where $|d|$ is the number of domains.
    \item \textbf{VREx} penalizes the variance of risk across domains We adapt VREx by using the empirical risk from Equation (\ref{eq:general_emp_risk}):
        \begin{equation} \label{eq:vrex_pen_val} \tag{VREx}
            P(f) = \mathsf{Var}_{D^d\in D}\big(R^d(f)\big),
        \end{equation}
    where $\mathsf{Var}$ is the variance taken across domains.
    \item \textbf{GroupDRO} performs importance weighting of the domains when calculating the empirical risk. We adapt the domain weighting parameter $\mathbf q_d$ using the empirical risk from Equation (\ref{eq:general_emp_risk}):
        \begin{equation} \label{eq:gdro_weights} \tag{GroupDRO}
            \mathbf q_d = \frac{\mathbf q'_d e^{R^d(f)}}{\sum_{D^d\in D}\mathbf q'_d(f)},
        \end{equation}
    where $\mathbf q'_d$ is the domain weights from the previous iteration.
    \item \textbf{IB-ERM} penalizes the variance of representation within domains. Consider a representation map $\Phi$ (that transforms inputs $\mathbf X$ as $\Phi(\mathbf X)$) and a linear classifier $w$ such that our predictor $f$ is defined as $w\cdot \Phi$. We define the IB-ERM penalty as:
        \begin{equation} \label{eq:ib_erm_pen_val} \tag{IB-ERM}
            P(f) = \frac{1}{|d|}\sum_{D^d\in D} \mathsf{Var}_{(\mathbf X, \mathbf Y)\in D^d}\big(\Phi(\mathbf X)\big),
        \end{equation}
    where $\mathsf{Var}$ is the variance is taken across samples of a domain.
    \item \textbf{SD} penalizes the squared l2 norm of the logits of the predictor $f$:
        \begin{equation} \label{eq:sd_pen_val} \tag{SD}
            P(f) = \frac{1}{n^d} \sum_{(\mathbf X^i, \mathbf Y^i)\in D} \frac{1}{|S_p^{d,i}|}\sum_{t\in S_p^{d,i}}  \|f(X_{1:t}^i)\|^2,
        \end{equation}
    where $n^d$ is the number of samples from domain $d$ in the dataset $D$.
\end{itemize}

\section{Measuring the impact of distribution shifts} \label{appendix:measuring_ood_gap}
We use the \textit{generalization gap} to empirically measure the impact of the distribution shifts on the performance of models. It measures the drop in performance between data drawn In-Distribution (ID) and Out-of-Distribution (OOD), where the former is independent and identically distributed (i.i.d.) to the training distribution and the later is not. \footnote{Some restriction with respect to the training distribution is implied, see Section~\ref{sect:prob_formulation_static}.} However, the generalization gap can be a misleading measure as it does not intrinsically indicate attainable performance gains. We show an example of unattainable performance gains later in this section. In this work, we do our best to measure an achievable performance gap for our dataset, i.e., an upper bound to the achievable performance on unseen domains. In this section we give details on of the generalization gaps described in Table~\ref{table:gen_gap} are obtained for domain generalization and subpopulation shift datasets.

\subsection{Generalization gap for domain generalization} \label{appendix:gen_gap_domain_generalization}
Given a set of training domains $D^\mathsf{train}$ and a test domain $D^\mathsf{test}$, we measure the OOD performance by training a model on the training domains $D^\mathsf{train}$ with ERM and measure the performance of this model on the test domain $D^\mathsf{test}$. The ID performance can be measured in multiple ways. \citet{wilds} provides multiple definitions for it:
\begin{itemize}
    \item \textbf{Train-to-train} Performance of a model on $D^\mathsf{train}$ when trained on $D^\mathsf{train}$
    \item \textbf{Mixed-to-test} Performance of a model on $D^\mathsf{test}$ when trained on a mixture of $D^\mathsf{train}$ and $D^\mathsf{test}$.
    \item \textbf{Test-to-test} Performance of a model on $D^\mathsf{test}$ when trained on $D^\mathsf{test}$
\end{itemize} 
We use the mixed-to-test measure for ID performance, because test-to-test and train-to-train can lead to erroneous measures leading to an inflated generalization gap that is unattainable in reality. To illustrate this problem, consider the generalization gap obtained with the train-to-train ID performance on the Spurious-Fourier dataset.

\begin{example}[Unattainable performance gap] \label{example:unattainable_gap}
    Performing ERM on the training domains $D^d|_{d\in\{80\%,\,90\%\}}$ will lead to a model relying on the spurious features to make predictions, as they are a stronger predictor of the label ($85\%$) than the invariant features ($75\%$). Thus, the model will achieve $85\%$ accuracy on data sampled ID to domains $d\in\{80\%,\,90\%\}$, but only achieve $10\%$ accuracy on the test domain $D^{10\%}$. Comparing the ID and OOD performance would lead to a generalization gap of $75\%$. However, this gap is misleading as a model could never achieve $85\%$ accuracy on the test domain because the strongest invariant predictor can only achieve $75\%$. A similar case can be made for the test-to-test measure of ID performance, where the generalization gap lead to an unattainable performance on the test domain.
\end{example}

Instead, consider the generalization gap obtained with the mixed-to-test ID performance for the same dataset.
\begin{example}[Attainable performance gap] \label{example:attainable_gap}
    Performing ERM on the training domains $D^d|_{d\in\{10\%,\,80\%,\,90\%\}}$ will lead to a predictor that relies on the invariant features, as they are a stronger predictor of the label ($75\%$) than the spurious features ($60\%$). Therefore, the ID performance will be $75\%$, and the OOD performance will be $10\%$, leading to a generalization gap of $65\%$. This gap is a much more significant measure of the upper bound of the performance than the original definition.
\end{example}

To summarize, we compute the generalization gap for domain generalization datasets as follows. Given a set of training domains $D^\mathsf{train}$ and a test domain $D^\mathsf{test}$, we first measure the \textit{OOD performance} by training a model on the training domains $D^\mathsf{train}$ with ERM and measure the performance of this model on the test domain $D^\mathsf{test}$. Second, we measure the \textit{ID performance} by training a model on all domains $D=\cup_{d=\{\mathsf{train}, \mathsf{test}\}} D^d$ and evaluate the model on the test domain $D^\mathsf{test}$. The generalization gap for that test domain is then defined as the difference between the ID and OOD performance. That process can then be repeated for all domains in the dataset and we average the performance.

\subsection{Generalization gap for subpopulation shifts}
In subpopulation shift datasets, we measure the OOD performance as the worst domain performance, and the ID performance as the train-to-train performance, i.e., the average domain performance. We recognize that this measure is not a perfect because of similar arguments made in Section~\ref{appendix:gen_gap_domain_generalization}, e.g., one domain might be much more difficult that the others and thus might be impossible to achieve the level of average performance on it. However, we argue that it is reasonable to consider the gap between the average domain performance and the worst domain performance as attainable. As a sanity check, one could verify that the average performance is achievable for a domain $d$ by doing a test-to-test style measure. We leave this to future work.

To summarize, we compute the generalization gap for subpopulation shifts datasets as follows. Given a set of training domains $D^\mathsf{train}$. we measure the ID performance by training a model on all domains, and measure the average performance across domains. We define the OOD performance as the performance of the model on the worst domain in $D^\mathsf{train}$. The generalization gap for that dataset is then defined as the different between the ID and OOD performance.

\section{Evaluation framework workflow} \label{appendix:framework}

In this section of the appendix, we detail the methodology employed to evaluate the performance of OOD generalization algorithms on our datasets. We follow the workflow used by \citet{domainbed} in their DomainBed testbed and adapt the framework to time series tasks.

\subsection{Reported performance}
We detail in this section the performance measure used for the different datasets in the WOODS benchmark. 

\paragraph{Synthetic challenge datasets}
Spurious-Fourier, TCMNIST-Source and TCMNIST-Time were formulated to address specific OOD generalization challenges in time series; thus we only investigate the training and testing domain configuration of interest, i.e., $D^\mathsf{train} = D^d|_{d\in\{80\%,\,90\%\}}$ and $D^\mathsf{test} = D^{10\%}$. With this domain configuration, we perform a hyperparameter sweep, the model selection (see Section~\ref{appendix:model_selection}) and report the performance of the chosen model on the 10\% domain.

\paragraph{Real-world domain generalization datasets} 
We report the performance of an OOD generalization algorithm with a domain cross-validation measure as follows. For every domain in a dataset, we perform a hyperparameter sweep with that domain held out from training. After this hyperparameter search, we perform the model selection associated with the dataset (see Section~\ref{appendix:model_selection}) and report the performance of the chosen model on the held out test set. We then report the average performance across domains.

\paragraph{Real-world subpopulation shift datasets}
We report the performance of an OOD generalization algorithm with the worst domain performance. We perform a hyperparameter sweep with all domains in the training dataset $D^\mathsf{train}$. After this search, we perform model selection (see Section~\ref{appendix:model_selection}) and report the worst domain performance.

\subsection{Systematic framework}
\paragraph{Hyperparameter search}
All hyperparameter searches in this work use random searches~\cite{randomsearch} over the hyperparameter distribution spaces defined in Table~\ref{table:training_hyperparameters} and Table~\ref{table:algorithm_hyperparameters}. We train 20 models using randomly sampled hyperparameter configurations. We then select the best performing model of those 20 configurations using different validation sets definitions, see Appendix~\ref{appendix:model_selection}. 

\paragraph{Statistically relevant}
We repeat each hyperparameter search three times to obtain statistically relevant results. This reduces the probability that some algorithm samples a lucky configuration of hyperparameters. All the results reported in this work are averaged over those three trials with a different seed. We also provide the estimated standard deviation of those averaged results.

\paragraph{Reducing bias}
The search range is an important topic when discussing the fairness of this evaluation strategy. Having reasonable hyperparameter distributions for sampling in the random search is essential to remaining fair between the algorithms and reducing the induced bias in the results. Defining a narrow hyperparameter distribution for which one knows the algorithm performs very well on a dataset or test domain leads to a bias of the evaluation due to queries of the test domain through human intervention. This bias could lead to algorithms getting better results by increasing the chance of the random search finding a good value. When defining the hyperparameter range, one should define a range wide enough as to cover at least the relevant search space for this hyperparameter. In this work we use ranges that accurately reflects the range of useful hyperparameters values, see Table~\ref{table:algorithm_hyperparameters}.

\begin{table}[h]
    \centering
    \caption{Distributions of training hyperparameters for random search} 
    \begin{tabular}{llll}
        \toprule
        \textbf{Dataset} & \textbf{Hyperparameter} & \textbf{Random distribution}\\
        \midrule
        \multirow{4}{*}{Spurious-Fourier}   & learning rate & $10^{\mathsf{Uniform}(-4.5, -2.5)}$ \\
                                            & batch size    & $2^{\mathsf{Uniform}(3, 9)}$ \\
                                            & class balance & True \\
        \midrule
        \multirow{4}{*}{TCMNIST-Source}     & learning rate & $10^{\mathsf{Uniform}(-4.5, -2.5)}$ \\
                                            & batch size    & $2^{\mathsf{Uniform}(3, 9)}$ \\
                                            & class balance & True \\
        \midrule
        \multirow{4}{*}{TCMNIST-Time}       & learning rate & $10^{\mathsf{Uniform}(-4.5, -2.5)}$ \\
                                            & batch size    & $2^{\mathsf{Uniform}(3, 9)}$ \\
                                            & class balance & True \\
        \midrule
        \multirow{4}{*}{CAP}                & learning rate & $10^{\mathsf{Uniform}(-5, -3)}$ \\
                                            & batch size    & $2^{\mathsf{Uniform}(3, 4)}$ \\
                                            & class balance & True \\
        \midrule
        \multirow{4}{*}{SEDFx}              & learning rate & $10^{\mathsf{Uniform}(-5, -3)}$ \\
                                            & batch size    & $2^{\mathsf{Uniform}(3, 4)}$ \\
                                            & class balance & True \\
        \midrule
        \multirow{4}{*}{PCL}                & learning rate & $10^{\mathsf{Uniform}(-5, -3)}$ \\
                                            & batch size    & $2^{\mathsf{Uniform}(3, 5)}$ \\
                                            & class balance & True \\
        \midrule
        \multirow{4}{*}{LSA64}              & learning rate & $10^{\mathsf{Uniform}(-5, -3)}$ \\
                                            & batch size    & $2^{\mathsf{Uniform}(3, 4)}$ \\
                                            & class balance & True \\
        \midrule
        \multirow{4}{*}{HHAR}               & learning rate & $10^{\mathsf{Uniform}(-4, -2)}$ \\
                                            & batch size    & $2^{\mathsf{Uniform}(3, 4)}$ \\
                                            & class balance & True \\
        \midrule
        \multirow{4}{*}{AusElec}            & learning rate & $10^{\mathsf{Uniform}(-5, -3)}$ \\
                                            & batch size    & $2^{\mathsf{Uniform}(3, 5)}$ \\
                                            & class balance & True \\

        \midrule
        \multirow{4}{*}{IEMOCAP}            & learning rate & $10^{\mathsf{Uniform}(-5, -3)}$ \\
                                            & batch size    & $2^{\mathsf{Uniform}(1, 4)}$ \\
                                            & class balance & True \\
        \bottomrule
    \end{tabular}
    \label{table:training_hyperparameters}
\end{table}

\begin{table}[h]
    \centering
    \caption{Distributions of algorithm hyperparameters for random search} 
    \begin{tabular}{llll}
        \toprule
        \textbf{Dataset} & \textbf{Hyperparameter} & \textbf{Random distribution}\\
        \midrule
        \multirow{2}{*}{Invariant Risk Minimization}    & penalty weight        & $10^{\mathsf{Uniform}(-1, 5)}$ \\
                                                        & annealing iterations  & $\mathsf{Uniform}(0, 2000)$ \\
        \midrule
        \multirow{2}{*}{Variational REx}                & penalty weight        & $10^{\mathsf{Uniform}(-1, 5)}$ \\
                                                        & annealing iterations  & $\mathsf{Uniform}(0, 2000)$ \\
        \midrule
        GroupDRO                                          & $\eta$     & $10^{\mathsf{Uniform}(-3, -1)}$ \\
        \midrule
        IB-ERM                                          & penalty weight     & $10^{\mathsf{Uniform}(-3, 0)}$ \\
        \midrule
        Spectral Decoupling                             & penalty weight        & $10^{\mathsf{Uniform}(-5, -1)}$ \\
        \midrule
        CAD                               & Lambda     & {Choice($[10^{-4}, 10^{-3}, 10^{-2}, 10^{-1}, 1, 10^{1}, 10^{2}]$)} \\
                                                        & Temperature  & {Choice($[0.05, 0.1]$)} \\
        \midrule
        CondCAD                               & lambda       & {Choice($[10^{-4}, 10^{-3}, 10^{-2}, 10^{-1}, 1, 10^{1}, 10^{2}]$)} \\
                                                        & {temperature}  & {Choice($[0.05, 0.1]$)} \\
        \midrule
        {Transfer}                          & lambda        & $10^{\mathsf{Uniform}(-2, 1)}$ \\
                                                        & delta        & $\mathsf{Uniform}(0.1, 3.0)$ \\
                                                        & adv lr         & $10^{\mathsf{Uniform}(-4.5,-2.5)}$ \\
                                                        & adv steps  & Choice($[1,2,5]$) \\
        \bottomrule
    \end{tabular}
    \label{table:algorithm_hyperparameters}
\end{table}

\section{Model Selection} \label{appendix:model_selection}

Section~\ref{appendix:framework} detailed the hyperparameter search and uncertainty estimation used in this framework. In this section, we detail the model selection strategy used in hyperparameter sweeps to determine the model to evaluate on the test domain.

\subsection{Model selection for domain generalization}
A fundamental restriction in domain generalization is that the training procedure does not have access to the test domains during training. As a result, the challenge of OOD generalization is not only to create models that generalize to the test domains but also to select the right models without having access to the test domains. Many model selection strategies were proposed~\cite{domainbed, ood-bench, wilds}, the simplest of which is \textit{Train-domain validation}.

\paragraph{Train-domain validation}
    We split the training domains into training and validation sets. The training split of the training domain is used to train the model. We choose the model that gets the best average validation performance across training domains. We report the performance of the chosen model on the testing domains.

However, tackling both problems of creating and finding invariant models at the same time might be a very difficult research endeavor. Instead, we can first start by narrowing the scope and only focus on creating invariant models. For this purpose, we relax the fundamental restriction and allow the queries of the test domain to obtain some signal on the absolute performance of an algorithm. Although querying the test domain can never be considered a valid model selection strategy in practical scenarios, the results can be very insightful when evaluating the behavior of an algorithm. \citet{domainbed} formulated \textit{Test-domain validation} that queries the test domains to perform model selection.

\paragraph{Test-domain validation}
    We split the test domains into testing and validation sets. Models are trained for a fixed number of training steps on the training domains. We choose the model with the best performance on the validation set of the test domains. However, we only consider the final checkpoint of the model after a fixed number of steps, effectively disallowing early stopping. We report the performance of the chosen model on the testing set of test domains.

Test-domain validation has proven to be a very useful measure of performance for algorithms on synthetic datasets driven primarily by correlation shift~\cite{ood-bench}, e.g., CMNIST. In such datasets, simple spurious features highly correlated with the label create shortcuts in the data that model leverage to minimize the empirical risk quickly (e.g., cow or camel classification problem). As a result, these shortcuts lead to very high training domain performance and very low test domain performance early in training. Consequently, any model selection criteria that rely on performance on data drawn i.i.d. to the training distribution is a poor way to investigate the performance of an algorithm because there is a bias of model selection towards early training correlation. Thus, by disallowing early stopping, we obtain an insightful measure to investigate the absolute performance of an algorithm.

On the other hand, Test-domain validation is ill-equipped to provide meaningful measures of performance with other kinds of datasets. For example, Test-domain validation is not an insightful measure of performance when dealing with real-world datasets. The reason is that we often do not know beforehand the number of training steps required for a given set of hyperparameters such that a model will finish the learning of the task. Therefore, we train models past the point of overfitting and pick the model with the highest validation performance. This renders the last checkpoint in training suboptimal for generalization performance, both ID and OOD, and leads to an uninformative measure of the generalization performance.

We introduce a more pragmatic model selection method that queries the test domain for real-world datasets to resolve this problem: \textit{Oracle train-domain validation}.

\paragraph{Oracle train-domain validation}
    We split the training domains into training and ID validation splits. We also split the test domains into testing and OOD validation splits. For every model training run, we choose the early stopped model that performs best on the ID validation split. Among all early stopped model of the sweep, we then choose the model that performs the best in the OOD validation split. Notice that this model selection method has the same number of queries of the test domain as the test-domain validation, i.e., one query per training run. 

In light of the discussion of this section, we use two different sets of model selection methods for the two different types of datasets in WOODS: Synthetic challenge and real-world datasets. We use \textbf{train-domain validation} and \textbf{test-domain validation} for our synthetic challenge datasets driven by correlation shift. We use \textbf{train-domain validation} and \textbf{oracle train-domain validation} for our real-world datasets which are likely driven by other kinds of shifts.

\subsection{Model selection for subpopulation shifts}
Model selection in subpopulation shift dataset is a much simpler endeavor because access to domains is not restricted. We define the two model selection strategies for our real-world datasets as follows.

\paragraph{Average domain validation}
We split all domains into training, validation and testing splits. The training splits of that dataset is used to train the model. We choose the model that gets the best average validation performance on all domains. We report the worst testing split performance of the chosen model.

\paragraph{Worst-domain validation}
We split all domains into training, validation and testing splits. The training splits of that dataset is used to train the model. We choose the model that gets the best worst domain validation performance. We report the worst testing split performance of the chosen model.

\end{document}